\documentclass[10pt,twocolumn,letterpaper]{article}

\usepackage[pagenumbers]{cvpr} 

\usepackage{booktabs}
\usepackage{multirow}
\usepackage{graphicx}
\usepackage{amsmath}
\usepackage{amssymb}
\usepackage{microtype}

\usepackage{marvosym}

\usepackage{fontawesome5}   
\usepackage{twemojis}       

\definecolor{cvprblue}{rgb}{0.21,0.49,0.74}
\usepackage[pagebackref,breaklinks,colorlinks,allcolors=cvprblue]{hyperref}

\def\confName{CVPR}
\def\confYear{2027}

\title{CellPlex: Benchmarking Single-Cell Spatial Proteomics Prediction from H\&E}

\author{%
\textbf{Zijun Gao}$^{1,2}$ \qquad
\textbf{Yijiang Chen}$^{2\,\text{\Letter}}$ \qquad
\textbf{Xiangde Luo}$^{2\,\text{\Letter}}$ \qquad
\textbf{Pheng-Ann Heng}$^{1}$\\[2pt]
{\normalsize $^{1}$The Chinese University of Hong Kong \qquad $^{2}$Stanford University School of Medicine}\\[8pt]
{\normalsize
\begin{tabular}{@{}l@{\hspace{0.2em}}c@{\hspace{0.2em}}l@{}}
  Code: & \faGithub       & \href{https://github.com/zjgao02/CellPlex}{\texttt{zjgao02/CellPlex}} \\[0.2em]
  Model: & \twemoji{1f917} & \href{https://huggingface.co/gaozijun/CellPlex}{\texttt{gaozijun/CellPlex}}
\end{tabular}}
}
\newcommand{\correspondence}{{\renewcommand{\thefootnote}{\Letter}%
  \footnotetext{Corresponding authors:\newline \texttt{yijiangc@stanford.edu}, \texttt{luoxd96@stanford.edu}}}}

\begin{document}
\maketitle
\correspondence
\begin{abstract}
Spatial proteomics measures tens of proteins in every cell of a tissue section, but its cost confines it to small research cohorts, whereas hematoxylin and eosin (H\&E) slides exist for nearly every patient.
Inferring single-cell spatial proteomics from H\&E would bring its analyses to routine pathology, and a growing number of virtual staining models pursue this goal.
Each is evaluated on its own cohort, panel and pixel-level metric, and released models transfer poorly to new cohorts, leaving open which approach to build on and which proteins H\&E can reveal.
We introduce CellPlex, a benchmark that poses the task at the level of the cell: seven cohorts with same-section H\&E and multiplexed immunofluorescence, from colorectal and breast cancer to lymph-node metastases, bone marrow and glioma, with 717 samples, 44.8 million cells and 149 proteins, processed by one pipeline into per-cell targets.
Pathology foundation models (PFMs) are adapted with one cell-level protocol, specialist models are fine-tuned under a matched budget, and all are scored on the same cells.
Adapted PFMs are the most accurate models on every cohort, ahead of the specialists even after fine-tuning, but keep less than half of their accuracy on cohorts they were not trained on.
Which proteins can be predicted depends on the tissue, not on how many cells were measured for them.
By fixing the cells, targets and splits on which predictions are judged, CellPlex sets a common standard for inferring costly molecular measurements from routine H\&E and shows how far H\&E can stand in for spatial proteomics.
We will release the data, splits, evaluation code and baselines as a testbed for developing and comparing future pathology foundation models and virtual staining methods.
\end{abstract}

\begin{figure*}[t]
  \centering
  \includegraphics[width=\linewidth]{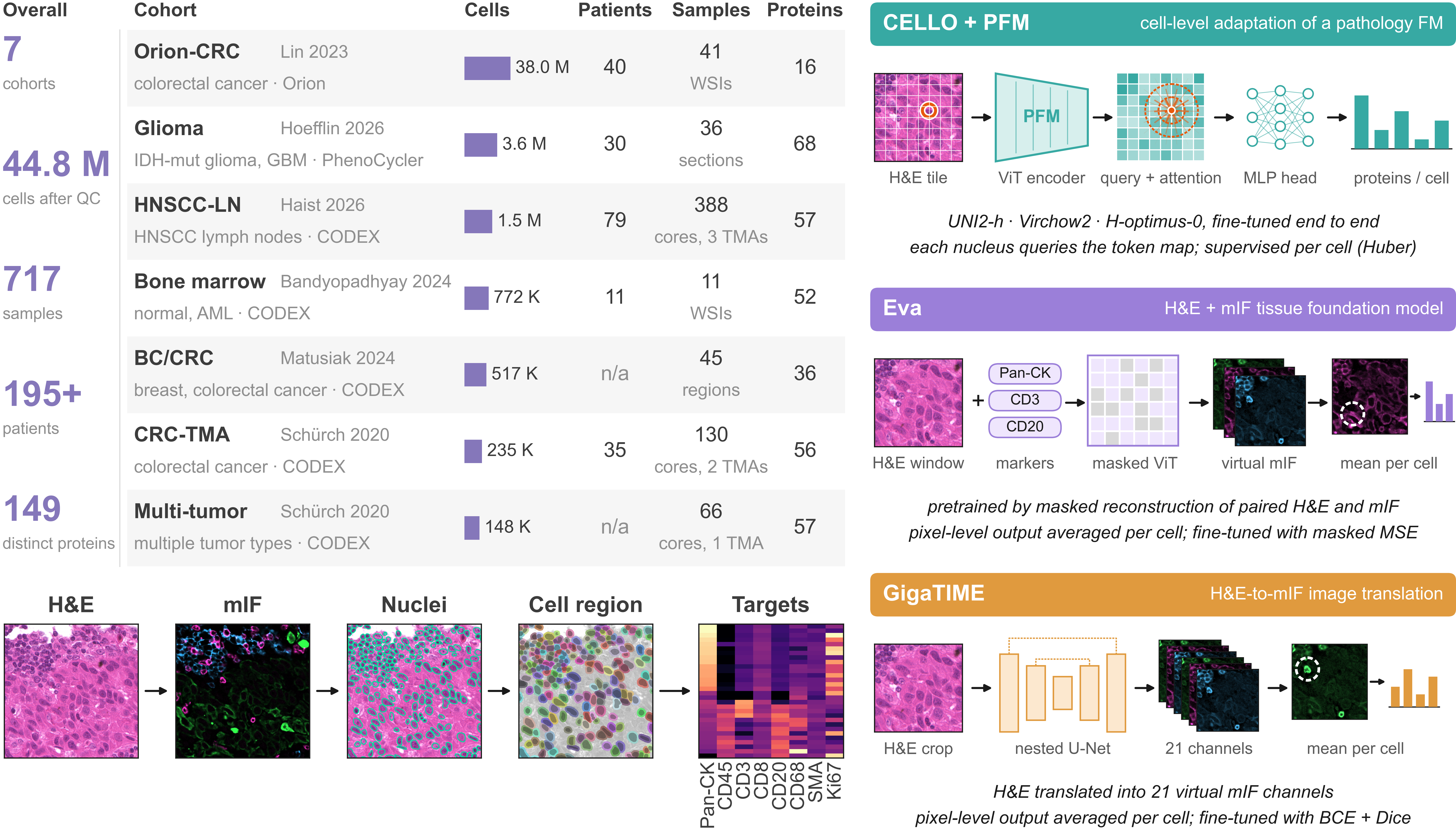}
  \caption{Overview of CellPlex. Left: the seven cohorts with their cells after quality control, patients, samples and proteins, and the construction of per-cell targets, in which nuclei segmented on the H\&E are grown into cell regions and the registered multiplexed image is averaged over each region. Right: the evaluated models, pathology foundation models adapted with a cell-level protocol and two specialist models, Eva and GigaTIME, whose pixel-level outputs are averaged over the same cell regions.}
  \label{fig:overview}
\end{figure*}

\section{Introduction}
\label{sec:intro}

Tissue function and disease arise from the molecular states of individual cells and from how these cells are arranged.
Spatial proteomics resolves both: imaging mass cytometry~\cite{giesen2014highly}, CODEX/PhenoCycler~\cite{goltsev2018deep,black2021codex}, MIBI~\cite{keren2019mibi} and Orion~\cite{lin2023high} measure tens of proteins in every cell of an intact tissue section, exposing immune phenotypes, checkpoint expression, stromal composition and cellular neighborhoods~\cite{schurch2020coordinated,karimi2024method}.
These readouts inform the study of tumor microenvironments and therapy response and are moving toward clinical decision support~\cite{quail2024revolutionizing,liu2026spatial}.
Their cost, throughput and consumption of tissue, however, keep them confined to research cohorts: the largest public collection of spatial proteomics, spora, covers 5,477 patients~\cite{von2026spora}.
Hematoxylin and eosin (H\&E) slides, in contrast, are produced for nearly every patient, and pathology foundation models (PFMs) have been pretrained on hundreds of thousands to millions of them~\cite{chen2024towards,zimmermann2024virchow2,saillard2024hoptimus0}.

H\&E morphology carries much of the information that multiplexed panels make explicit, so inferring single-cell spatial proteomics from H\&E would extend these analyses to routine pathology and its archives.
A growing number of models pursue this goal by learning from paired H\&E and multiplexed images: image-to-image translators that predict a fixed panel~\cite{burlingame2020shift,andani2025histopathology,wu2025rosie,valanarasu2026multimodal}, predictors built on PFMs~\cite{balezo2026miphei,li2026ai}, and foundation models for spatial proteomics that take H\&E as further input channels and, once fine-tuned for the task, predict any requested marker from H\&E alone~\cite{liu2025modeling}.

The field cannot yet tell which of these approaches to build on, because each model has been evaluated on its own terms.
The models are trained on different and mostly small corpora (GigaTIME on 21 whole slides and HistoPlexer on 336 regions, each from a single cohort, ROSIE on 1,342 cores~\cite{von2026spora}, MIPHEI-ViT on Orion-CRC~\cite{balezo2026miphei}), predict different panels, and are scored mainly on held-out data from their own sources.
In the spora benchmark~\cite{von2026spora}, GigaTIME, HistoPlexer and ROSIE were applied as released to four cohorts with registered H\&E and scored pixel by pixel against the measured images.
Averaged over the markers that all three predict, the Pearson correlation stayed below 0.11 for every model and cohort.
The second metric, structural similarity (SSIM), ranked the models differently: on the intestine cohort of Hickey \etal~\cite{hickey2023organization}, HistoPlexer obtained the highest SSIM (0.60) although its predictions were uncorrelated with the measurements (Pearson 0.004), so the two metrics did not even agree on which model was best.
Because none of the three models was fine-tuned on these cohorts, the comparison cannot tell whether the low scores reflect the limits of H\&E, of the models or of their training data.
Benchmarks of PFMs do not fill this gap: they target slide- and patch-level tasks and, for molecular targets, spot-level transcriptomics probed with frozen features~\cite{jaume2024hest}, where each spot averages the expression of several neighbouring cells, often of different types.

We introduce CellPlex, a benchmark that poses the task at the level of the cell (\cref{fig:overview}).
It assembles seven cohorts in which H\&E and multiplexed immunofluorescence were acquired on the same tissue section, a pairing that only five of the 35 cohorts in spora provide~\cite{von2026spora}.
The cohorts span colorectal and breast cancer, a multi-tumor array, lymph-node metastases of head-and-neck cancer, bone marrow and glioma on two imaging platforms, with 717 samples, 44.8 million cells and 149 proteins.
Turning these releases into a benchmark is less a matter of collecting data than of making the task well posed.
Nuclei are segmented on the H\&E, so that every target belongs to a cell visible at inference time; the multiplexed image is registered to the H\&E and every sample is checked for residual misalignment; and the target of a cell is the mean level of each protein over its nucleus and a 2\,$\mu$m rim.
Test cells come from patients or samples held out from training, and every model is scored on the same cells by correlation and structural similarity.

CellPlex compares models under controlled conditions.
Every PFM is turned into a cell-level predictor by the same protocol, which samples its token map at each nucleus and refines the feature by attention over the surrounding tile; the architecture follows CELLO~\cite{gao2026towards}, and both it and the training recipe are fixed so that the PFM is the only variable.
A linear probe on frozen features measures what the PFM provides before adaptation.
Two specialist models, Eva~\cite{liu2025modeling} and GigaTIME~\cite{valanarasu2026multimodal}, are evaluated as released and after fine-tuning on every cohort with their own objectives and a training budget matched to the PFMs.
Finally, models trained on one cohort are applied to every other cohort to measure transfer.

Evaluating UNI2-h~\cite{chen2024towards}, Virchow2~\cite{zimmermann2024virchow2} and H-optimus-0~\cite{saillard2024hoptimus0} in this way, we find:
\begin{itemize}
    \item \textbf{Adapted PFMs are the strongest models.} CELLO with H-optimus-0 reaches a mean PCC of 0.532 over the seven cohorts, against 0.460 for fine-tuned Eva and 0.344 for fine-tuned GigaTIME, and is the most accurate model on every cohort; a linear probe on its frozen features reaches 0.446.
    \item \textbf{Transfer across cohorts remains open.} Trained on one cohort and applied without further training to another, CELLO with H-optimus-0 reaches a mean PCC of 0.22 over the proteins shared by the two panels, against 0.52 when trained on the target cohort itself and 0.21 for Eva used as released. Transfer works only between closely related cohorts: trained on CRC-TMA, the model reaches 0.39 on the Multi-tumor array, which comes from the same study and panel, against 0.44 when trained on it, whereas on glioma it reaches at most 0.15 when trained on another cohort, against 0.46 on the same proteins when trained on glioma.
    \item \textbf{Which proteins can be predicted depends on the tissue.} The same protein can be easy in one cohort and hard in another: with CELLO (H-optimus-0), CD3 reaches 0.66 PCC on Orion-CRC but 0.20 on glioma, and Ki67 0.74 on bone marrow but 0.30 on the Multi-tumor array. Accuracy is unrelated to how many cells a protein was measured in (Spearman $\rho = -0.11$ over 149 proteins) and is higher for proteins whose measured levels are spatially coherent: 0.57 in the upper two quintiles of spatial autocorrelation (Moran's $I$), against 0.46--0.47 in the lower two.
\end{itemize}
These results place the remaining difficulty in transfer across tissues and in proteins whose levels vary from cell to cell rather than with tissue architecture.
We will release the processed data, splits, evaluation code and baselines of CellPlex as a testbed for future PFMs and virtual staining methods.

\section{Related Work}
\label{sec:related}

\paragraph{Spatial proteomics prediction from H\&E.}
Models that predict spatial proteomics from H\&E, often called virtual staining, learn from sections in which H\&E and a multiplexed protein image were acquired together and registered.
The first models were image-to-image translators trained on one cohort for a fixed panel: SHIFT~\cite{burlingame2020shift} used a conditional GAN to predict a few immunofluorescence markers, and HistoPlexer~\cite{andani2025histopathology} extended this design to an imaging mass cytometry panel, trained on 336 regions of metastatic melanoma.
ROSIE~\cite{wu2025rosie} scaled the training data to more than 1,300 H\&E--CODEX samples from over a dozen tissues and predicts 50 proteins, and GigaTIME~\cite{valanarasu2026multimodal}, a nested U-Net that predicts 21 immunofluorescence channels, was trained on 21 whole slides and applied to the archival slides of more than 14,000 patients to build a virtual population.
More recent models replace the convolutional encoder with a PFM.
MIPHEI-ViT~\cite{balezo2026miphei} attaches a pixel decoder to H-optimus-0 and predicts 16 channels of Orion-CRC, and HEX~\cite{li2026ai} partially fine-tunes a PFM to regress the mean level of 40 proteins over tiles of about 50\,$\mu$m, trained on ten whole slides of lung cancer and used to predict prognosis and immunotherapy response.
These models also differ in how they are evaluated: most by pixel-level correlation or structural similarity on held-out data from their own source, MIPHEI-ViT by classifying marker-positive cells, and HEX by correlation over tiles and by downstream clinical prediction, so their reported accuracies are not comparable.

\paragraph{Spatial proteomics foundation models.}
A second line of work learns representations from the multiplexed images themselves.
VirTues~\cite{wenckstern2026virtual} and KRONOS~\cite{shaban2025foundation} are pretrained on multiplexed images, with marker embeddings that let one model accept different panels; they take the measured protein channels as input, so they analyse spatial proteomics where it has been acquired but do not extend it to slides with H\&E alone.
Eva~\cite{liu2025modeling} follows a similar channel-agnostic design and adds the matched H\&E of each region as three further input channels of a masked autoencoder, pretrained on more than 4,000 regions and 64 million cells.
Predicting proteins from H\&E is not part of this pretraining: it required a separate fine-tuning stage in which every multiplexed channel is masked, and the resulting model, released by its authors for virtual staining and evaluated here, reached a median per-channel correlation of 0.22 in their evaluation, on par with ROSIE.
Eva's data are also concentrated on one platform family and on cancer, a limitation its authors note: 88\% of its curated images come from CODEX or PhenoCycler and 96\% from cancers.
The spora resource~\cite{von2026spora} harmonizes 35 public cohorts for training such models, and its H\&E task scores three released translators zero-shot and per pixel (\cref{sec:intro}).

\paragraph{Spatial transcriptomics prediction from H\&E.}
Gene expression prediction from H\&E began with sequencing-based spatial transcriptomics, whose 55-$\mu$m Visium spots each average several cells: models regress spot expression from patches~\cite{he2020integrating}, retrieve it through contrastive embeddings~\cite{xie2023spatially}, add the context of neighbouring spots and of the slide~\cite{chung2024accurate,nonchev2025deepspot}, or pretrain image and expression encoders jointly~\cite{chen2025visual,xiang2026multimodal}.
Imaging-based platforms have since moved the task to single cells: DeepSpot2Cell~\cite{nonchev2025deepspot2cell} is trained on spot-level targets, and CELLO~\cite{gao2026towards}, whose architecture we reuse, on single-cell targets, sampling a PFM's token map at each cell and refining it by distance-aware cross-attention.
Transcripts are, however, an indirect readout of cell function: mRNA levels explain only part of the variation in protein levels~\cite{liu2016dependency}, and the markers that type cells and grade their states, such as CD3, CD8, Ki67 and PD-L1, are proteins.

\paragraph{Pathology foundation models.}
PFMs such as UNI and UNI2-h~\cite{chen2024towards}, Virchow2~\cite{zimmermann2024virchow2}, Prov-GigaPath~\cite{xu2024whole}, H-optimus-0~\cite{saillard2024hoptimus0} and CONCH~\cite{lu2024visual} are vision transformers~\cite{dosovitskiy2020image} pretrained on large H\&E collections, mostly with DINOv2-style objectives~\cite{oquab2023dinov2}, and are compared on slide- and patch-level tasks; HEST-1k~\cite{jaume2024hest} added spot-level gene expression probed with frozen features, and none of these benchmarks asks for per-cell outputs or proteins.

\section{CellPlex}
\label{sec:benchmark}

\subsection{Task}
\label{sec:task}
Given an H\&E image and the nuclei segmented on it, the task is to predict, for every nucleus $i$, a vector $\hat{\mathbf{y}}_i \in \mathbb{R}^{P}$ of the levels of the $P$ proteins in the cohort's panel.
The target is $y_{i,p} = \log\!\big(1 + \bar{x}_{i,p}\big)$, where $\bar{x}_{i,p}$ is the mean intensity of channel $p$ of the registered multiplexed image over the region $R_i$ formed by the nucleus and a 2\,$\mu$m rim around it, grown without entering neighbouring cells.
Using nuclei segmented on the H\&E, rather than cells segmented on the multiplexed image, makes the task well posed at inference time, when only the H\&E is available, and ties every target to an object the model can see.
Models receive the H\&E at its native resolution and the nucleus positions, never the multiplexed image.

\subsection{Cohorts}
\label{sec:cohorts}
CellPlex combines seven public cohorts with same-section H\&E and multiplexed immunofluorescence; \cref{tab:cohorts} in the supplementary material lists their platform, resolution, size, panel and split unit.
\textbf{Orion-CRC}~\cite{lin2023high} contains whole slides of 40 colorectal cancer patients imaged with the Orion platform, with 16 proteins after removing the nuclear stain and autofluorescence channels; at 38 million cells it is by far the largest cohort.
\textbf{CRC-TMA}~\cite{schurch2020coordinated} consists of two CODEX tissue microarrays of 35 colorectal cancer patients with four cores each and 56 proteins.
The same study provides a \textbf{Multi-tumor} array stained with a closely related 57-protein panel.
\textbf{BC/CRC}~\cite{matusiak2024spatially} contains 45 CODEX regions of breast (16) and colorectal (29) cancer with a 36-protein panel focused on macrophage and stromal markers.
\textbf{HNSCC-LN}~\cite{haist2026lymph} is the discovery cohort of a study of lymph-node metastasis in head-and-neck squamous cell carcinoma, with 388 cores from 79 patients and 57 proteins.
\textbf{Bone marrow}~\cite{bandyopadhyay2024mapping} contains whole-slide CODEX images of bone marrow biopsies measured with two panels that share 52 proteins; the only biopsy from acute myeloid leukemia fails quality control (\cref{sec:processing}), leaving 11 samples.
\textbf{Glioma}~\cite{hoefflin2026spatial} contains whole sections of IDH-mutant astrocytoma, oligodendroglioma and glioblastoma from 30 patients, imaged with PhenoCycler-Fusion, the commercial form of CODEX, and stained with H\&E afterwards, with 68 proteins.
The cohorts differ in tissue, platform, resolution (0.25--0.51\,$\mu$m/px), sample format and panel, and together cover 149 distinct proteins, from lineage markers (CD45, Pan-cytokeratin, CD3, CD20) to functional and state markers (Ki67, PD-1, PD-L1, Granzyme B, HIF1$\alpha$).

\subsection{Processing}
\label{sec:processing}
\paragraph{Nuclei.}
Nuclei are segmented on the H\&E of every cohort with CellViT++~\cite{horst2024cellvit,horst2026cellvit++}, run at about 0.25\,$\mu$m/px; images at other resolutions are resampled for segmentation and the contours mapped back to the native pixel grid, which all later steps use.

\paragraph{Registration.}
On the CODEX arrays the H\&E is scanned on the same microscope grid as the multiplexed image, up to a translation of a few micrometers.
We estimate this translation per core by cross-correlating the segmented nuclei with the nuclear stain channel within $\pm$40 pixels.
For BC/CRC the released H\&E is already on the CODEX grid, and for Orion-CRC and bone marrow the released multiplexed images are registered to the H\&E.
The glioma H\&E was scanned on another microscope at 0.25\,$\mu$m/px; we register the multiplexed image to it by a coarse rigid search on tissue masks, an affine alignment of the hematoxylin channel to the nuclear stain and a smoothed field of local displacements estimated on 260\,$\mu$m tiles.

\paragraph{Ground truth and quality control.}
Each nucleus is grown by 2\,$\mu$m without entering its neighbours, and the mean intensity of every protein channel over the region gives the target of \cref{sec:task}.
As a registration and staining check, we compute for every sample the fraction of cells whose nuclear stain is brighter inside the nucleus than in the surrounding rim; a misregistered sample loses this contrast, and samples below 0.8 are excluded.
This removes 10 CRC-TMA cores, one BC/CRC region, one HNSCC core, one bone marrow sample and one glioma section; one further HNSCC core without clinical annotation is not used.

\subsection{Splits and metrics}
\label{sec:splits}
\paragraph{Splits.}
All cohorts except Orion-CRC and glioma use five folds.
Folds are formed by patient for HNSCC-LN (stratified by nodal status) and bone marrow, by core for CRC-TMA and Multi-tumor, and by region for BC/CRC (stratified by tissue type).
For fold $k$, the units of fold $k$ form the test set, the first half of the units of fold $k{+}1$ the validation set used for model selection, and the rest the training set.
The two cohorts with the largest tissue area use one fixed patient-level split: Orion-CRC, whose slides hold on average almost one million cells, with 28 training, 4 validation and 8 test patients, and glioma, stratified by tumor type, with 21 training, 3 validation and 6 test patients (9 sections).

\paragraph{Metric.}
For every test sample and protein we compute the Pearson correlation coefficient (PCC) between predicted and measured levels across all of the sample's cells, and average over the proteins of the panel to obtain the sample's score.
As a second metric we compute the structural similarity index (SSIM) over the cells of a sample, following its use for spot-level expression maps~\cite{li2022benchmarking}: the predicted and measured levels $x$ and $y$ of a protein are each min--max scaled to $[0,1]$ over the sample's cells, and $\mathrm{SSIM} = \frac{(2\mu_x\mu_y + C_1)(2\sigma_{xy} + C_2)}{(\mu_x^2 + \mu_y^2 + C_1)(\sigma_x^2 + \sigma_y^2 + C_2)}$ with $C_1 = 0.01^2$ and $C_2 = 0.03^2$.
Unlike PCC, SSIM also penalizes a mismatch in the mean and spread of the scaled levels.
We report the mean of the sample scores over the test samples of all folds (\cref{tab:main}), with standard errors in \cref{tab:main_se} of the supplementary material, and compare models with two-sided Wilcoxon signed-rank tests on paired sample scores.
Correlation within a sample asks whether a model ranks the cells of one tissue correctly, which is what downstream analyses of cell types, states and neighborhoods require, and it is insensitive to the per-sample intensity offsets that staining and imaging introduce.

\section{Methods}
\label{sec:method}

\subsection{Adapting PFMs to single cells with CELLO}
\label{sec:protocol}
Every PFM is turned into a cell-level predictor with the architecture of CELLO~\cite{gao2026towards} and one training recipe (\cref{sec:supp_training}), so that the PFM is the only component that changes between runs.
Each H\&E image is cut into non-overlapping $L \times L$ tiles at the cohort's native resolution, with $L = 224$.
The PFM encodes a tile into a class token $\mathbf{z}_{\mathrm{cls}} \in \mathbb{R}^d$ and a token map $\mathbf{T} \in \mathbb{R}^{L_g \times L_g \times d}$, where $L_g = L/p$ for patch size $p$.
For a cell centered at $\mathbf{s}_i$ in the tile, the query $\mathbf{q}_i = \operatorname{GridSample}(\mathbf{T}, \mathbf{s}_i)$ is sampled bilinearly from the token map and attends to all tokens of the tile through a cross-attention with 2-D rotary position embeddings~\cite{su2024roformer} and the distance-decay bias
\begin{equation}
  b_{i,m} = -\frac{\|\mathbf{s}_i - \mathbf{c}_m\|^2}{2p^2},
\end{equation}
where $\mathbf{c}_m$ is the center of token $m$.
The layer-normalized output is fused with $\mathbf{z}_{\mathrm{cls}}$ by a residual attention layer and decoded by a two-layer MLP with ReLU output into the $P$ proteins of the panel.
Training minimizes the Huber loss over cells plus 0.5 times the Huber loss of a second head on $\mathbf{z}_{\mathrm{cls}}$ that predicts the summed target of the tile's cells, and the PFM is fine-tuned together with the new layers.
The linear probe samples the frozen token map at each nucleus in the same way and maps it to the panel by ridge regression, with the regularization strength chosen on the validation samples.
We evaluate UNI2-h~\cite{chen2024towards} and Virchow2~\cite{zimmermann2024virchow2}, both ViT-H/14, and H-optimus-0~\cite{saillard2024hoptimus0}, a ViT-g/14.

\subsection{Specialist models}
\label{sec:specialists}
We compare the adapted PFMs with two published H\&E-to-multiplex models, run with their released weights.
\textbf{Eva}~\cite{liu2025modeling} is a foundation model for spatial proteomics that takes H\&E as three further input channels; we use the checkpoint its authors fine-tuned and released for H\&E-to-multiplex prediction, which takes 224-pixel H\&E windows at 0.5\,$\mu$m/px and predicts any requested marker through a marker embedding; markers absent from its vocabulary are requested through the text embedding of their gene~\cite{chen2024genept}, so that the full panel of every cohort is predicted.
\textbf{GigaTIME}~\cite{valanarasu2026multimodal} is a nested U-Net that maps H\&E at 0.25\,$\mu$m/px to 21 immunofluorescence channels; zero-shot, it is scored only on the panel proteins that match one of its channels (7 to 14 per cohort).
For both models, the prediction for a cell is the mean predicted intensity over the same region $R_i$ that defines the target, so that all models are scored on identical cells and targets.

\paragraph{Fine-tuning.}
To separate the value of a specialist's architecture and pretraining from its exposure to the test domain, we also fine-tune both models on the training folds of every cohort with their own objectives.
Targets are the registered multiplexed images, scaled per sample and channel to $[0, 1]$ between the 0.5th and 99.5th percentiles over tissue.
Eva is trained on all parameters with the objective of its authors' H\&E-to-multiplex fine-tuning, the mean-squared error of the multiplexed channels predicted from H\&E alone.
GigaTIME receives one output channel per panel protein, initialized from the matching pretrained channel where one exists, and is trained with its binary cross-entropy and Dice losses.
The training budget of both is capped at the tissue area that the adapted PFMs see on the same cohort, and the checkpoint with the best validation score is tested (\cref{sec:supp_training}).

\section{Experiments}
\label{sec:results}

\begin{table*}[t]
\centering
\resizebox{\linewidth}{!}{\setlength{\aboverulesep}{0pt}\setlength{\belowrulesep}{0pt}\renewcommand{\arraystretch}{1.2}
\begin{tabular}{l|cc|cc|cc|cc|cc|cc|cc|cc}
\toprule
 & \multicolumn{2}{c|}{Orion-CRC} & \multicolumn{2}{c|}{CRC-TMA} & \multicolumn{2}{c|}{Multi-tumor} & \multicolumn{2}{c|}{BC/CRC} & \multicolumn{2}{c|}{HNSCC-LN} & \multicolumn{2}{c|}{Bone marrow} & \multicolumn{2}{c|}{Glioma} & \multicolumn{2}{c}{Avg} \\
\cmidrule(lr){2-3} \cmidrule(lr){4-5} \cmidrule(lr){6-7} \cmidrule(lr){8-9} \cmidrule(lr){10-11} \cmidrule(lr){12-13} \cmidrule(lr){14-15} \cmidrule(lr){16-17}
Model & PCC & SSIM & PCC & SSIM & PCC & SSIM & PCC & SSIM & PCC & SSIM & PCC & SSIM & PCC & SSIM & PCC & SSIM \\
\midrule
\textit{Specialist models} & & & & & & & & & & & & & & & & \\
Eva, zero-shot & 0.381 & 0.159 & 0.190 & 0.133 & 0.105 & 0.082 & 0.188 & 0.161 & 0.142 & 0.107 & 0.168 & 0.104 & 0.118 & 0.076 & 0.185 & 0.117 \\
Eva, fine-tuned & 0.500 & 0.322 & 0.491 & 0.340 & 0.419 & 0.298 & 0.456 & 0.397 & 0.441 & 0.295 & 0.548 & 0.411 & 0.365 & 0.239 & 0.460 & 0.329 \\
GigaTIME, zero-shot & 0.173 & 0.059 & 0.104 & 0.082 & 0.006 & 0.018 & 0.053 & 0.050 & 0.078 & 0.036 & $-$0.055 & $-$0.014 & 0.024 & 0.031 & 0.055 & 0.038 \\
GigaTIME, fine-tuned & 0.426 & 0.248 & 0.356 & 0.237 & 0.297 & 0.217 & 0.317 & 0.257 & 0.333 & 0.222 & 0.358 & 0.311 & 0.323 & 0.207 & 0.344 & 0.243 \\
\midrule
\textit{Linear probe} & & & & & & & & & & & & & & & & \\
UNI2-h & 0.593 & 0.434 & 0.426 & 0.378 & 0.332 & 0.293 & 0.414 & 0.336 & 0.380 & 0.354 & 0.531 & 0.498 & 0.402 & 0.402 & 0.440 & 0.385 \\
Virchow2 & 0.585 & 0.449 & 0.417 & 0.369 & 0.342 & 0.299 & 0.404 & 0.326 & 0.366 & 0.346 & 0.525 & 0.486 & 0.408 & 0.411 & 0.435 & 0.384 \\
H-optimus-0 & 0.590 & 0.435 & 0.432 & 0.384 & 0.339 & 0.297 & 0.421 & 0.339 & 0.389 & 0.360 & 0.538 & 0.506 & 0.413 & 0.413 & 0.446 & 0.391 \\
\midrule
\textit{CELLO} & & & & & & & & & & & & & & & & \\
UNI2-h & 0.615 & 0.471 & 0.512 & 0.479 & 0.426 & 0.393 & 0.514 & 0.456 & 0.466 & 0.440 & 0.607 & 0.585 & 0.492 & 0.498 & 0.519 & 0.475 \\
Virchow2 & 0.595 & 0.458 & 0.508 & 0.477 & 0.431 & 0.398 & 0.514 & 0.457 & 0.455 & 0.433 & 0.602 & 0.581 & 0.480 & 0.487 & 0.512 & 0.470 \\
H-optimus-0 & \textbf{0.620} & \textbf{0.487} & \textbf{0.530} & \textbf{0.499} & \textbf{0.443} & \textbf{0.411} & \textbf{0.531} & \textbf{0.476} & \textbf{0.479} & \textbf{0.455} & \textbf{0.620} & \textbf{0.595} & \textbf{0.503} & \textbf{0.505} & \textbf{0.532} & \textbf{0.490} \\
\bottomrule
\end{tabular}
}
\caption{Single-cell protein prediction on CellPlex. For each cohort, Pearson correlation (PCC) and cell-level structural similarity (SSIM) between predicted and measured protein levels. Each score is averaged over the proteins of a test sample and then over the test samples of a cohort; Avg: mean over the seven cohorts; standard errors over test samples are in \cref{tab:main_se}. \textbf{Bold}: best per column. Zero-shot GigaTIME is scored only on the panel proteins matching one of its 21 channels (7--14 per cohort).}
\label{tab:main}
\end{table*}

\subsection{CELLO-adapted PFMs versus specialist models}
\label{sec:main}
\Cref{tab:main} reports every model on every cohort, and \crefrange{tab:prot_orion}{tab:prot_glioma} in the supplementary material on every protein.
CELLO with H-optimus-0 is the most accurate model on all seven cohorts by both PCC and SSIM, with a PCC of 0.44 (Multi-tumor) to 0.62 (Orion-CRC and bone marrow).
It is more accurate than fine-tuned Eva on 573 of the 657 test samples and significantly so on every cohort ($p \leq 0.008$, \cref{tab:paired} in the supplementary material), with a mean PCC of 0.532 against 0.460; fine-tuned Eva comes closest on Multi-tumor (0.419 against 0.443) and is furthest behind on glioma (0.365 against 0.503).
Fine-tuned GigaTIME reaches 0.30 to 0.43 and is below CELLO with H-optimus-0 on 656 of the 657 test samples.
The CELLO adaptation, designed for single-cell transcriptomics, thus carries over to proteins and gives a strong baseline for new methods.

\paragraph{Linear probe.}
The frozen features of a PFM already carry much of this signal.
A linear probe on H-optimus-0 reaches a mean PCC of 0.446, close to fine-tuned Eva (0.460), is ahead of it on Orion-CRC (0.590 against 0.500) and glioma (0.413 against 0.365), and has a higher mean SSIM (0.391 against 0.329).
Adapting the PFM with CELLO adds 0.03 (Orion-CRC) to 0.11 (BC/CRC) PCC over the linear probe and is better on 646 of the 657 test samples.
Frozen PFM features thus already encode much of the protein signal, and adapting them to single cells adds a consistent gain; a linear probe is a cheap reference point for new methods.

\paragraph{Zero-shot and fine-tuned specialists.}
Used as released, both specialists transfer poorly.
Zero-shot Eva reaches 0.381 on Orion-CRC but only 0.105 to 0.190 on the six CODEX cohorts, and zero-shot GigaTIME stays at or below 0.173 even when scored only on the proteins it was trained to predict, with a negative correlation on bone marrow.
Fine-tuning raises Eva by 0.12 to 0.38 PCC, which suggests that much of its zero-shot gap reflects the shift in tissue, staining, imaging platform and panel between its training data and a new cohort.
On six of the seven cohorts this gain exceeds the remaining gap to CELLO with H-optimus-0 (0.28 against 0.07 on average), so training on paired data from the target cohort matters more than the choice of starting model.

\begin{figure*}[t]
  \centering
  \includegraphics[width=\linewidth]{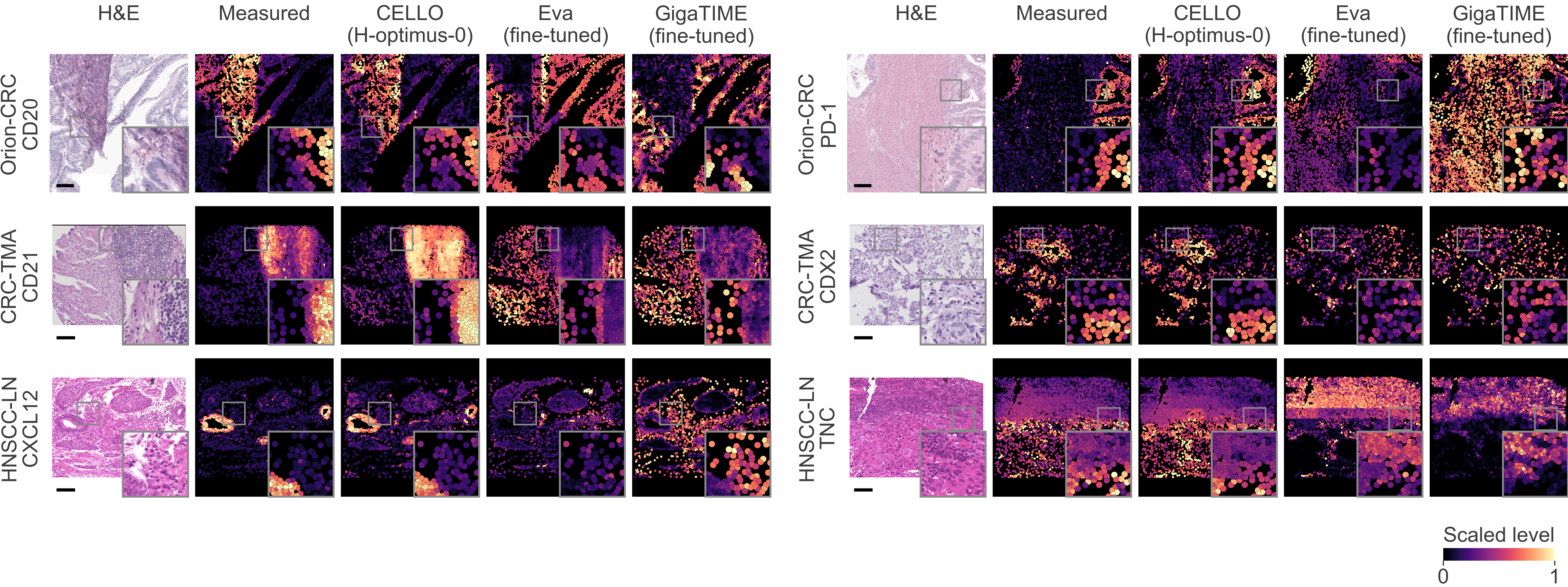}
  \caption{Protein maps on held-out samples of three cohorts, two per cohort (four more cohorts in \cref{fig:maps_supp}). Each case shows the H\&E of one view and its cells colored by one protein, as measured and as predicted by CELLO (H-optimus-0) and by fine-tuned Eva and GigaTIME; the inset magnifies the boxed region. Every panel is scaled to the 1st--99th percentile of its cells in the view; scale bar, 100\,$\mu$m.}
  \label{fig:maps}
\end{figure*}

\subsection{Comparing PFMs}
\label{sec:pfms}
With CELLO, H-optimus-0 is the most accurate of the three PFMs on every cohort (\cref{tab:main}).
It scores higher than UNI2-h on 67\% (glioma) to 100\% (bone marrow) of the test samples of a cohort, significantly on every cohort except glioma (6 of 9 sections, $p = 0.074$), and higher than Virchow2 on every cohort ($p \leq 0.008$, \cref{tab:paired}).
Its margins are 0.005 to 0.018 PCC over UNI2-h and 0.012 to 0.025 over Virchow2.
These differences are small next to the gain of CELLO over the linear probe (0.03 to 0.11) but consistent across cohorts, which suggests that CellPlex can separate encoders of similar accuracy.

\subsection{Transfer across cohorts}
\label{sec:transfer}
To measure transfer, the fold-0 CELLO model of each cohort is applied without further training to every other cohort, with the H\&E resampled to the resolution of the training cohort, and scored on the proteins shared by the two panels (\cref{tab:cross} in the supplementary material).
Over the 42 pairs, CELLO with H-optimus-0 reaches a mean PCC of 0.22, against 0.52 for the target cohort's own models on the same proteins and cells; the ratio of the two, averaged over pairs, is 41\% (\cref{sec:supp_transfer}).
Eva used as released reaches 0.21 on the same proteins, and the transferred model is ahead of it on only 20 of the 42 pairs.
Transfer succeeds only between closely related cohorts.
Trained on CRC-TMA, the model reaches 0.39 on the Multi-tumor array, which comes from the same study and panel, against 0.44 in-domain; trained on BC/CRC, it reaches 0.47 on Orion-CRC, another colorectal cohort, against 0.61.
Glioma is the hardest target: trained on any other cohort, the model reaches at most 0.15 on it, against 0.46 in-domain.
UNI2-h transfers slightly better than H-optimus-0 (mean PCC 0.226 against 0.219, Virchow2 0.202).
The adapted models thus appear to learn largely cohort-specific mappings from H\&E to protein levels, and transfer to cohorts without paired data remains open.

\begin{figure*}[t]
  \centering
  \includegraphics[width=\linewidth]{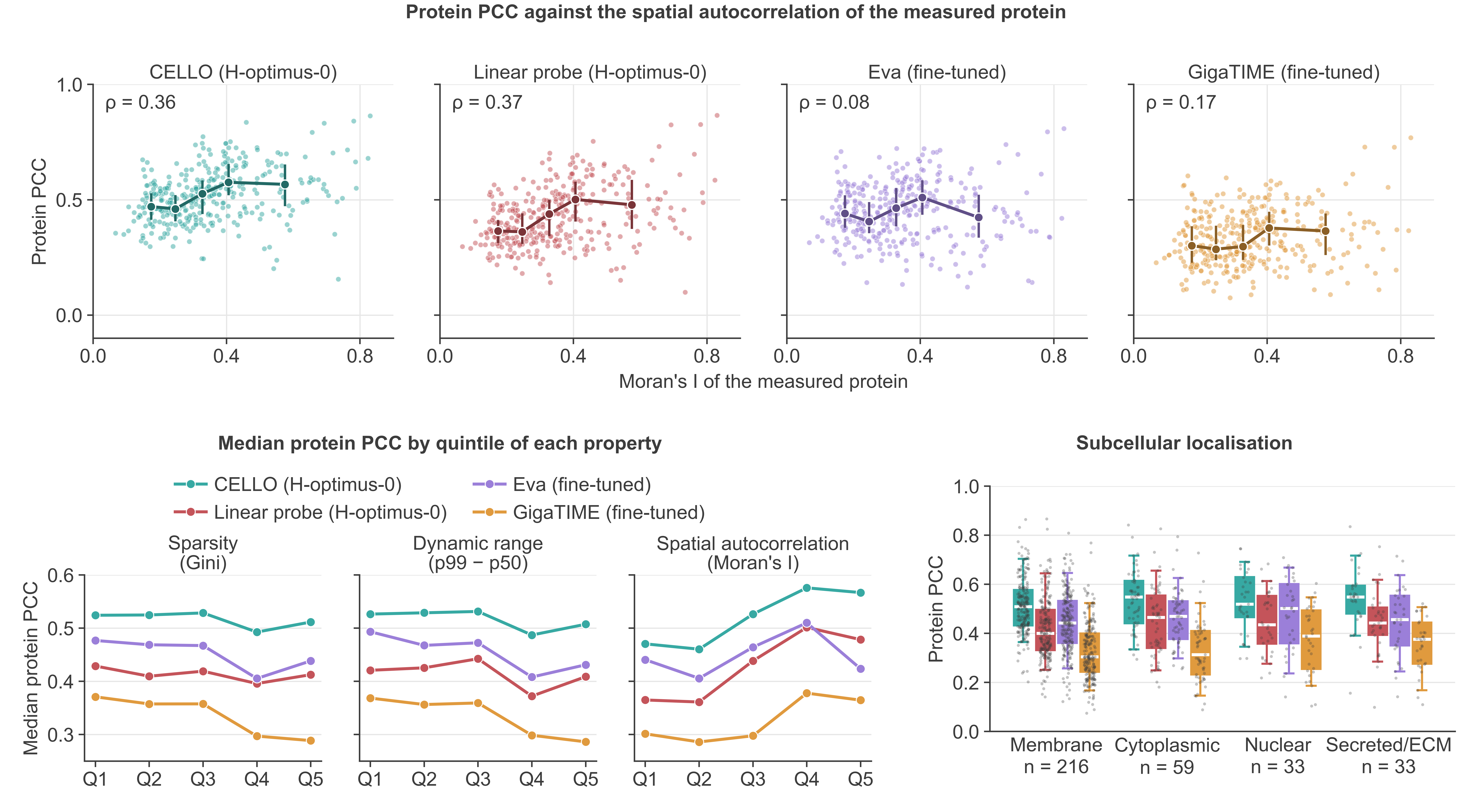}
  \caption{What makes a protein predictable, over the 341 cohort--protein pairs of the benchmark. Top: protein PCC against the spatial autocorrelation of the measured protein (Moran's $I$ over the 10 nearest cells); lines: median and interquartile range per quintile of Moran's $I$; $\rho$: Spearman correlation. Bottom left: median protein PCC per quintile of sparsity (Gini coefficient), dynamic range (99th minus 50th percentile of the measured level) and Moran's $I$. Bottom right: protein PCC by subcellular localization; boxes: interquartile range and median, whiskers: 5th--95th percentile; $n$: cohort--protein pairs.}
  \label{fig:props}
\end{figure*}

\subsection{Which proteins can be predicted}
\label{sec:proteins}
We score each protein of a cohort by its protein PCC, the mean of its PCC over the cohort's test samples (\cref{sec:supp_proteins} gives this and the other quantities of this section).
Within a cohort, CELLO with H-optimus-0 predicts some proteins far better than others, from a spread of 0.38 to 0.71 PCC on CRC-TMA to 0.16 to 0.84 on glioma (\cref{fig:protein_map} and \crefrange{tab:prot_orion}{tab:prot_glioma} in the supplementary material).
Which proteins are easy depends on the tissue.
For the 46 proteins measured in at least three cohorts, a two-way analysis of variance of protein PCC with protein and cohort as factors attributes 32\% of the variance to the protein, 21\% to the cohort and the remaining 47\% to their interaction, and the ranking of proteins agrees only partly between cohorts (Spearman $\rho$ over the shared proteins of two cohorts, from $-0.38$ to $0.69$).
CD3 reaches 0.66 on Orion-CRC but 0.20 on glioma, and Ki67 0.74 on bone marrow but 0.30 on Multi-tumor.
Accuracy is unrelated to how many cells a protein was measured in over all cohorts (Spearman $\rho = -0.11$ with the protein's mean PCC over its cohorts, 149 proteins, \cref{fig:protein_cells}).
Per-protein accuracy measured on one tissue may therefore not carry over to another, and its variation does not simply follow the amount of training data per protein.
At the protein level, CELLO with H-optimus-0 is more accurate than fine-tuned Eva on 297 of the 341 cohort--protein pairs and on most proteins of every cohort, so its advantage is not confined to a few proteins.

\Cref{fig:props} relates the protein PCC of every cohort--protein pair to properties of its measured levels, each computed on every test sample and averaged over the samples of the cohort.
Accuracy rises with spatial autocorrelation: the protein PCC of CELLO with H-optimus-0 has a Spearman correlation of 0.36 with the Moran's $I$ of the measured protein (Kruskal--Wallis $p = 7 \times 10^{-11}$ across the quintiles of the 341 pairs).
The linear probe follows the same trend ($\rho = 0.37$), whereas the accuracy of fine-tuned Eva and GigaTIME depends less on it ($\rho = 0.08$ and $0.17$).
Accordingly, the margin of the adapted PFM over fine-tuned Eva grows with Moran's $I$, from a median of 0.03 in the lowest quintile to 0.13 in the highest ($\rho = 0.52$).
Neither sparsity nor dynamic range changes the accuracy of the adapted PFM (Kruskal--Wallis $p \geq 0.3$), and neither does subcellular localization ($p = 0.2$): membrane, cytoplasmic, nuclear and secreted proteins reach median PCCs of 0.51 to 0.55.
In our cohorts, H\&E is thus most informative about proteins whose levels follow tissue architecture, irrespective of their subcellular localization, and proteins that vary between neighboring cells remain harder to predict.

\subsection{Qualitative results}
\label{sec:qualitative}
\Cref{fig:maps} shows two held-out samples for each of three cohorts, and \cref{fig:maps_supp} in the supplementary material two for each of the other four.
CELLO with H-optimus-0 reproduces the B-cell follicles of Orion-CRC (CD20) and of a CRC-TMA core (CD21), the clustered PD-1-high and CDX2-high cells of the same cohorts and the CXCL12-high rings and tenascin-C (TNC) pattern of HNSCC lymph-node cores, with their boundaries at the level of single nuclei ($r = 0.68$ to $0.83$, where $r$ is the PCC of the shown protein over the cells of the view, \cref{sec:supp_maps}).
The fine-tuned specialists instead place the signal in other structures (for instance, Eva spreads CD20 over the epithelium of Orion-CRC, GigaTIME predicts PD-1 across most of the Orion-CRC view, and both assign high CD21 outside the CRC-TMA follicle) and are at best uncorrelated with the measurements in these views ($r = -0.55$ to $0.00$).
Their errors concern which structures carry the signal rather than spatial resolution, which suggests that recognizing the context of each cell may matter more than producing sharper maps.

\section{Limitations and open problems}
\label{sec:limitations}
CellPlex leaves several problems open.
The first is transfer: models adapted on one cohort lose most of their accuracy on another, except between closely related cohorts (\cref{sec:transfer}).
Models that generalize across tissues, staining protocols, imaging platforms and panels remain to be developed, and the transfer protocol of CellPlex measures progress toward them.
The second is the panel: the seven panels share only part of their proteins, so a model trained on one cohort predicts only the proteins of its panel, and models that can pool cohorts with different panels or predict proteins not measured in a cohort are an open direction.
The third is the ceiling: the targets are multiplexed immunofluorescence measurements, whose noise bounds the score of any model, and part of the variation of a protein between cells may not be visible in H\&E at all; how close current models are to this ceiling, in particular for proteins that vary between neighboring cells, is not known.
Finally, seven cohorts on two imaging platforms cover only some tissues and diseases, and we score agreement with the measurement rather than the value of predicted proteins for downstream analyses; new cohorts, platforms, PFMs and cell-level heads can be added under the same pipeline and protocol.

\section{Conclusion}
\label{sec:conclusion}
CellPlex makes single-cell protein prediction from H\&E comparable across models, by fixing the cells, targets, splits and metrics on which seven same-section cohorts are judged.
Its baselines indicate that cell-level adaptation of general-purpose pathology foundation models is a strong starting point for the task, while paired data from the target cohort remains necessary for high accuracy.
They also mark where H\&E currently falls short: in tissues unlike those seen in training and for proteins whose levels do not follow tissue architecture.
We see three directions for future work: models trained on many paired cohorts that transfer across tissues and panels, models that capture protein variation between neighboring cells, and evaluations of predicted proteins in downstream tasks, from cell typing to outcome prediction in cohorts that have only H\&E.
We will release CellPlex as a common testbed for this work.

{
    \small
    \bibliographystyle{ieeenat_fullname}
    \bibliography{main}
}

\clearpage
\setcounter{section}{0}
\setcounter{table}{0}
\setcounter{figure}{0}
\renewcommand{\thesection}{S\arabic{section}}
\renewcommand{\thetable}{S\arabic{table}}
\renewcommand{\thefigure}{S\arabic{figure}}
\makeatletter
\twocolumn[{%
  \centering
  {\Large\textbf{\thetitle}\\
  \vspace{0.5em}Supplementary Material\\}
  \vspace{1.0em}
  \begin{minipage}{\textwidth}
  \def\@captype{table}%
\centering
\small
\setlength{\tabcolsep}{5pt}
\begin{tabular}{llllrrrrl}
\toprule
Cohort & Tissue & Platform & Format & $\mu$m/px & Samples & Cells & Proteins & Split unit \\
\midrule
Orion-CRC~\cite{lin2023high} & Colorectal cancer & Orion & WSI & 0.325 & 41 & 38.0\,M & 16 & patient (40) \\
CRC-TMA~\cite{schurch2020coordinated} & Colorectal cancer & CODEX & TMA & 0.377 & 130 & 234.5\,K & 56 & core \\
BC/CRC~\cite{matusiak2024spatially} & Breast, colorectal cancer & CODEX & regions & 0.507 & 45 & 516.9\,K & 36 & region \\
Multi-tumor~\cite{schurch2020coordinated} & Multiple tumor types & CODEX & TMA & 0.377 & 66 & 147.9\,K & 57 & core \\
HNSCC-LN~\cite{haist2026lymph} & HNSCC lymph nodes & CODEX & TMA & 0.377 & 388 & 1.47\,M & 57 & patient (79) \\
Bone marrow~\cite{bandyopadhyay2024mapping} & Bone marrow & CODEX & WSI & 0.249 & 11 & 772.4\,K & 52 & patient (11) \\
Glioma~\cite{hoefflin2026spatial} & Glioma & CODEX & WSI & 0.249 & 36 & 3.59\,M & 68 & patient (30) \\
\midrule
Total & & & & & 717 & 44.8\,M & 149 & \\
\bottomrule
\end{tabular}
\caption{Cohorts of CellPlex after quality control. H\&E and multiplexed immunofluorescence were acquired on the same tissue section in every cohort. Samples are whole-slide images (WSI), tissue-microarray cores or imaged regions; cells are nuclei segmented on the H\&E. The total protein count merges synonymous marker names across panels.}
\label{tab:cohorts}
  \end{minipage}
  \vspace{1.5em}
}]
\makeatother

\section{Cohorts}
\label{sec:supp_cohorts}
\Cref{tab:cohorts} summarizes the seven cohorts of CellPlex after quality control (\cref{sec:cohorts,sec:processing}).

\section{Standard errors and paired tests}
\label{sec:supp_se}
\Cref{tab:main_se} repeats the cohort scores of \cref{tab:main} with their standard errors over test samples; $n$ is the number of test samples, pooled over the five folds except for the fixed splits of Orion-CRC and glioma (\cref{sec:splits}).
\Cref{tab:paired} compares CELLO with H-optimus-0 sample by sample with the other two PFMs and with fine-tuned Eva (\cref{sec:main,sec:pfms}).

\begin{table*}[t]
\centering
\small
\resizebox{\linewidth}{!}{\begin{tabular}{lccccccc}
\toprule
Model & Orion-CRC & CRC-TMA & Multi-tumor & BC/CRC & HNSCC-LN & Bone marrow & Glioma \\
 & ($n = 8$) & ($n = 130$) & ($n = 66$) & ($n = 45$) & ($n = 388$) & ($n = 11$) & ($n = 9$) \\
\midrule
\multicolumn{8}{l}{\textit{PCC}} \\
Eva, zero-shot & 0.381\,$\pm$\,0.019 & 0.190\,$\pm$\,0.007 & 0.105\,$\pm$\,0.010 & 0.188\,$\pm$\,0.009 & 0.142\,$\pm$\,0.003 & 0.168\,$\pm$\,0.016 & 0.118\,$\pm$\,0.013 \\
Eva, fine-tuned & 0.500\,$\pm$\,0.026 & 0.491\,$\pm$\,0.008 & 0.419\,$\pm$\,0.011 & 0.456\,$\pm$\,0.013 & 0.441\,$\pm$\,0.003 & 0.548\,$\pm$\,0.024 & 0.365\,$\pm$\,0.021 \\
GigaTIME, zero-shot & 0.173\,$\pm$\,0.022 & 0.104\,$\pm$\,0.007 & 0.006\,$\pm$\,0.010 & 0.053\,$\pm$\,0.005 & 0.078\,$\pm$\,0.003 & $-$0.055\,$\pm$\,0.011 & 0.024\,$\pm$\,0.021 \\
GigaTIME, fine-tuned & 0.426\,$\pm$\,0.019 & 0.356\,$\pm$\,0.008 & 0.297\,$\pm$\,0.012 & 0.317\,$\pm$\,0.010 & 0.333\,$\pm$\,0.003 & 0.358\,$\pm$\,0.025 & 0.323\,$\pm$\,0.020 \\
Linear probe, UNI2-h & 0.593\,$\pm$\,0.018 & 0.426\,$\pm$\,0.008 & 0.332\,$\pm$\,0.010 & 0.414\,$\pm$\,0.008 & 0.380\,$\pm$\,0.004 & 0.531\,$\pm$\,0.018 & 0.402\,$\pm$\,0.014 \\
Linear probe, Virchow2 & 0.585\,$\pm$\,0.018 & 0.417\,$\pm$\,0.008 & 0.342\,$\pm$\,0.009 & 0.404\,$\pm$\,0.008 & 0.366\,$\pm$\,0.003 & 0.525\,$\pm$\,0.019 & 0.408\,$\pm$\,0.015 \\
Linear probe, H-optimus-0 & 0.590\,$\pm$\,0.019 & 0.432\,$\pm$\,0.008 & 0.339\,$\pm$\,0.010 & 0.421\,$\pm$\,0.009 & 0.389\,$\pm$\,0.003 & 0.538\,$\pm$\,0.019 & 0.413\,$\pm$\,0.014 \\
CELLO, UNI2-h & 0.615\,$\pm$\,0.024 & 0.512\,$\pm$\,0.009 & 0.426\,$\pm$\,0.013 & 0.514\,$\pm$\,0.011 & 0.466\,$\pm$\,0.003 & 0.607\,$\pm$\,0.021 & 0.492\,$\pm$\,0.014 \\
CELLO, Virchow2 & 0.595\,$\pm$\,0.028 & 0.508\,$\pm$\,0.009 & 0.431\,$\pm$\,0.013 & 0.514\,$\pm$\,0.010 & 0.455\,$\pm$\,0.003 & 0.602\,$\pm$\,0.023 & 0.480\,$\pm$\,0.017 \\
CELLO, H-optimus-0 & 0.620\,$\pm$\,0.025 & 0.530\,$\pm$\,0.009 & 0.443\,$\pm$\,0.013 & 0.531\,$\pm$\,0.011 & 0.479\,$\pm$\,0.003 & 0.620\,$\pm$\,0.020 & 0.503\,$\pm$\,0.012 \\
\midrule
\multicolumn{8}{l}{\textit{SSIM}} \\
Eva, zero-shot & 0.159\,$\pm$\,0.016 & 0.133\,$\pm$\,0.004 & 0.082\,$\pm$\,0.006 & 0.161\,$\pm$\,0.007 & 0.107\,$\pm$\,0.002 & 0.104\,$\pm$\,0.007 & 0.076\,$\pm$\,0.007 \\
Eva, fine-tuned & 0.322\,$\pm$\,0.027 & 0.340\,$\pm$\,0.005 & 0.298\,$\pm$\,0.007 & 0.397\,$\pm$\,0.011 & 0.295\,$\pm$\,0.002 & 0.411\,$\pm$\,0.015 & 0.239\,$\pm$\,0.010 \\
GigaTIME, zero-shot & 0.059\,$\pm$\,0.007 & 0.082\,$\pm$\,0.004 & 0.018\,$\pm$\,0.006 & 0.050\,$\pm$\,0.003 & 0.036\,$\pm$\,0.001 & $-$0.014\,$\pm$\,0.006 & 0.031\,$\pm$\,0.012 \\
GigaTIME, fine-tuned & 0.248\,$\pm$\,0.023 & 0.237\,$\pm$\,0.005 & 0.217\,$\pm$\,0.008 & 0.257\,$\pm$\,0.008 & 0.222\,$\pm$\,0.002 & 0.311\,$\pm$\,0.017 & 0.207\,$\pm$\,0.013 \\
Linear probe, UNI2-h & 0.434\,$\pm$\,0.029 & 0.378\,$\pm$\,0.007 & 0.293\,$\pm$\,0.010 & 0.336\,$\pm$\,0.009 & 0.354\,$\pm$\,0.003 & 0.498\,$\pm$\,0.019 & 0.402\,$\pm$\,0.010 \\
Linear probe, Virchow2 & 0.449\,$\pm$\,0.026 & 0.369\,$\pm$\,0.008 & 0.299\,$\pm$\,0.009 & 0.326\,$\pm$\,0.010 & 0.346\,$\pm$\,0.003 & 0.486\,$\pm$\,0.020 & 0.411\,$\pm$\,0.011 \\
Linear probe, H-optimus-0 & 0.435\,$\pm$\,0.026 & 0.384\,$\pm$\,0.007 & 0.297\,$\pm$\,0.010 & 0.339\,$\pm$\,0.010 & 0.360\,$\pm$\,0.003 & 0.506\,$\pm$\,0.021 & 0.413\,$\pm$\,0.010 \\
CELLO, UNI2-h & 0.471\,$\pm$\,0.038 & 0.479\,$\pm$\,0.009 & 0.393\,$\pm$\,0.013 & 0.456\,$\pm$\,0.013 & 0.440\,$\pm$\,0.003 & 0.585\,$\pm$\,0.023 & 0.498\,$\pm$\,0.013 \\
CELLO, Virchow2 & 0.458\,$\pm$\,0.038 & 0.477\,$\pm$\,0.009 & 0.398\,$\pm$\,0.014 & 0.457\,$\pm$\,0.012 & 0.433\,$\pm$\,0.003 & 0.581\,$\pm$\,0.024 & 0.487\,$\pm$\,0.016 \\
CELLO, H-optimus-0 & 0.487\,$\pm$\,0.038 & 0.499\,$\pm$\,0.009 & 0.411\,$\pm$\,0.013 & 0.476\,$\pm$\,0.013 & 0.455\,$\pm$\,0.003 & 0.595\,$\pm$\,0.022 & 0.505\,$\pm$\,0.012 \\
\bottomrule
\end{tabular}
}
\caption{Mean $\pm$ standard error over test samples of the PCC and SSIM in \cref{tab:main}. Each sample's score is its mean over the proteins of the panel; $n$: test samples.}
\label{tab:main_se}
\end{table*}

\begin{table}[t]
\centering
\small
\setlength{\tabcolsep}{3.5pt}
\resizebox{\linewidth}{!}{\begin{tabular}{lcccccc}
\toprule
 & \multicolumn{2}{c}{vs.\ UNI2-h} & \multicolumn{2}{c}{vs.\ Virchow2} & \multicolumn{2}{c}{vs.\ Eva (FT)} \\
\cmidrule(lr){2-3} \cmidrule(lr){4-5} \cmidrule(lr){6-7}
Cohort & wins & $p$ & wins & $p$ & wins & $p$ \\
\midrule
Orion-CRC & 7/8 & 0.016 & 8/8 & 0.008 & 8/8 & 0.008 \\
CRC-TMA & 117/130 & $<\!10^{-19}$ & 123/130 & $<\!10^{-21}$ & 116/130 & $<\!10^{-16}$ \\
BC/CRC & 40/45 & $<\!10^{-7}$ & 37/45 & $<\!10^{-6}$ & 44/45 & $<\!10^{-12}$ \\
Multi-tumor & 54/66 & $<\!10^{-6}$ & 46/66 & $<\!10^{-3}$ & 45/66 & 0.003 \\
HNSCC-LN & 330/388 & $<\!10^{-39}$ & 367/388 & $<\!10^{-58}$ & 340/388 & $<\!10^{-51}$ \\
Bone marrow & 11/11 & $<\!10^{-3}$ & 11/11 & $<\!10^{-3}$ & 11/11 & $<\!10^{-3}$ \\
Glioma & 6/9 & 0.074 & 8/9 & 0.008 & 9/9 & 0.004 \\
\bottomrule
\end{tabular}
}
\caption{Paired comparison of CELLO (H-optimus-0) with the other PFMs and with fine-tuned Eva: number of test samples on which H-optimus-0 scores higher, and two-sided Wilcoxon signed-rank $p$-value over test samples.}
\label{tab:paired}
\end{table}

\section{Training details}
\label{sec:supp_training}
\paragraph{Adapted PFMs.}
All runs use AdamW~\cite{loshchilov2017decoupled} with a learning rate of $2 \times 10^{-4}$ (weight decay 0.05) for the new layers and a peak of $2 \times 10^{-5}$ (weight decay 0.01) for the PFM, cosine schedules, batches of 64 tiles with at most 256 cells per tile, mixed precision, gradient clipping at 1, and flips with brightness and contrast jitter.
The PFM is frozen for the first 10\% of training, its learning rate is warmed up from $5 \times 10^{-6}$ over the next 30\%, and it is then fine-tuned with the rest of the model.
The distance-decay bias is clipped at $-20$.
Array and region cohorts are trained for 200 epochs with validation every 5 epochs; the whole-slide cohorts, which have far more tiles, are trained for 5 epochs (Orion-CRC and glioma) or 50 epochs (bone marrow) with validation after every epoch.
The checkpoint with the best validation score, the mean over validation samples of the sample's mean protein PCC, is tested.
Each run uses one GPU.

\paragraph{Linear probe.}
Token features are extracted once per cohort with the frozen PFM.
The ridge regression includes an intercept, and its regularization strength is a multiple of the mean eigenvalue of the centered feature covariance, chosen on a logarithmic grid by the validation score used for the adapted PFMs.

\paragraph{Specialist models.}
Eva receives the H\&E at 0.5\,$\mu$m/px as $1 - \mathrm{RGB}/255$ in its three H\&E channels, in 224-pixel windows, and is trained with AdamW at a learning rate of $7.1 \times 10^{-5}$ (weight decay 0.05), 300 warm-up steps, cosine decay, gradient clipping at 1 and an effective batch of 32 windows.
GigaTIME receives ImageNet-normalized RGB at 0.25\,$\mu$m/px in 512-pixel crops processed as four 256-pixel windows; panel proteins without a matching pretrained channel are initialized from the mean of its protein channels, and it is trained with AdamW at a learning rate of $1.5 \times 10^{-4}$ (weight decay $10^{-4}$), 200 warm-up steps, cosine decay, gradient clipping at 1 and batches of 64 crops.
Both losses are computed over tissue pixels.
The training budget is expressed in each model's own windows, and the checkpoint with the best of eight evenly spaced validations, scored as for the adapted PFMs, is tested.

\section{Transfer across cohorts}
\label{sec:supp_transfer}
\Cref{tab:cross} gives the PCC of CELLO with H-optimus-0 for every pair of training and target cohort (\cref{sec:transfer}).

For a training cohort $a$ and a target cohort $b$, the fold-0 model of $a$ predicts the proteins shared by the two panels on every test sample of $b$ (the test samples of all folds, or the fixed test set of Orion-CRC and glioma), from H\&E resampled to the resolution of $a$.
Proteins are matched by their canonical name, and the pairs share 8 to 47 proteins.
The cross score of the pair is the mean over these samples of the sample's mean PCC over the shared proteins.
The in-domain score is computed in the same way, on the same samples and proteins, from the predictions of $b$'s own out-of-fold models with the same PFM, and the retention of the pair is the ratio of the cross to the in-domain score.
The mean PCCs of \cref{sec:transfer} are means of these scores over the 42 pairs, and the 41\% is the mean retention.
Eva used as released is scored in the same way on the same samples and shared proteins, and a pair counts in favor of the transferred model when its cross score is higher than that of Eva.

\begin{table*}[t]
\centering
\small
\resizebox{\linewidth}{!}{\begin{tabular}{lccccccc}
\toprule
Trained on & Multi-tumor & CRC-TMA & BC/CRC & HNSCC-LN & Bone marrow & Glioma & Orion-CRC \\
\midrule
Multi-tumor & \textbf{0.44} & 0.40 (0.53) & 0.24 (0.55) & 0.21 (0.47) & 0.15 (0.59) & 0.02 (0.48) & 0.30 (0.60) \\
CRC-TMA & 0.39 (0.44) & \textbf{0.53} & 0.32 (0.54) & 0.24 (0.47) & 0.16 (0.59) & 0.04 (0.46) & 0.42 (0.60) \\
BC/CRC & 0.17 (0.44) & 0.31 (0.53) & \textbf{0.53} & 0.26 (0.49) & 0.28 (0.58) & 0.08 (0.48) & 0.47 (0.61) \\
HNSCC-LN & 0.18 (0.44) & 0.28 (0.53) & 0.34 (0.53) & \textbf{0.48} & 0.12 (0.59) & 0.07 (0.47) & 0.33 (0.60) \\
Bone marrow & 0.14 (0.41) & 0.19 (0.54) & 0.21 (0.53) & 0.17 (0.49) & \textbf{0.62} & 0.15 (0.46) & 0.26 (0.61) \\
Glioma & 0.10 (0.42) & 0.09 (0.53) & 0.12 (0.52) & 0.10 (0.48) & 0.26 (0.60) & \textbf{0.50} & 0.24 (0.60) \\
Orion-CRC & 0.18 (0.43) & 0.34 (0.54) & 0.34 (0.58) & 0.23 (0.48) & 0.18 (0.60) & 0.12 (0.51) & \textbf{0.62} \\
\bottomrule
\end{tabular}
}
\caption{Transfer of CELLO with H-optimus-0 across cohorts. Rows: training cohort, whose fold-0 model is applied without further training; columns: target cohort. Off-diagonal: PCC on the proteins shared by the two panels, with the target cohort's own models on the same proteins and cells in parentheses. Diagonal (bold): the target cohort's own models on its full panel (\cref{tab:main}).}
\label{tab:cross}
\end{table*}

\section{Per-protein accuracy}
\label{sec:supp_proteins}
\begin{figure*}[t]
  \centering
  \includegraphics[width=\linewidth]{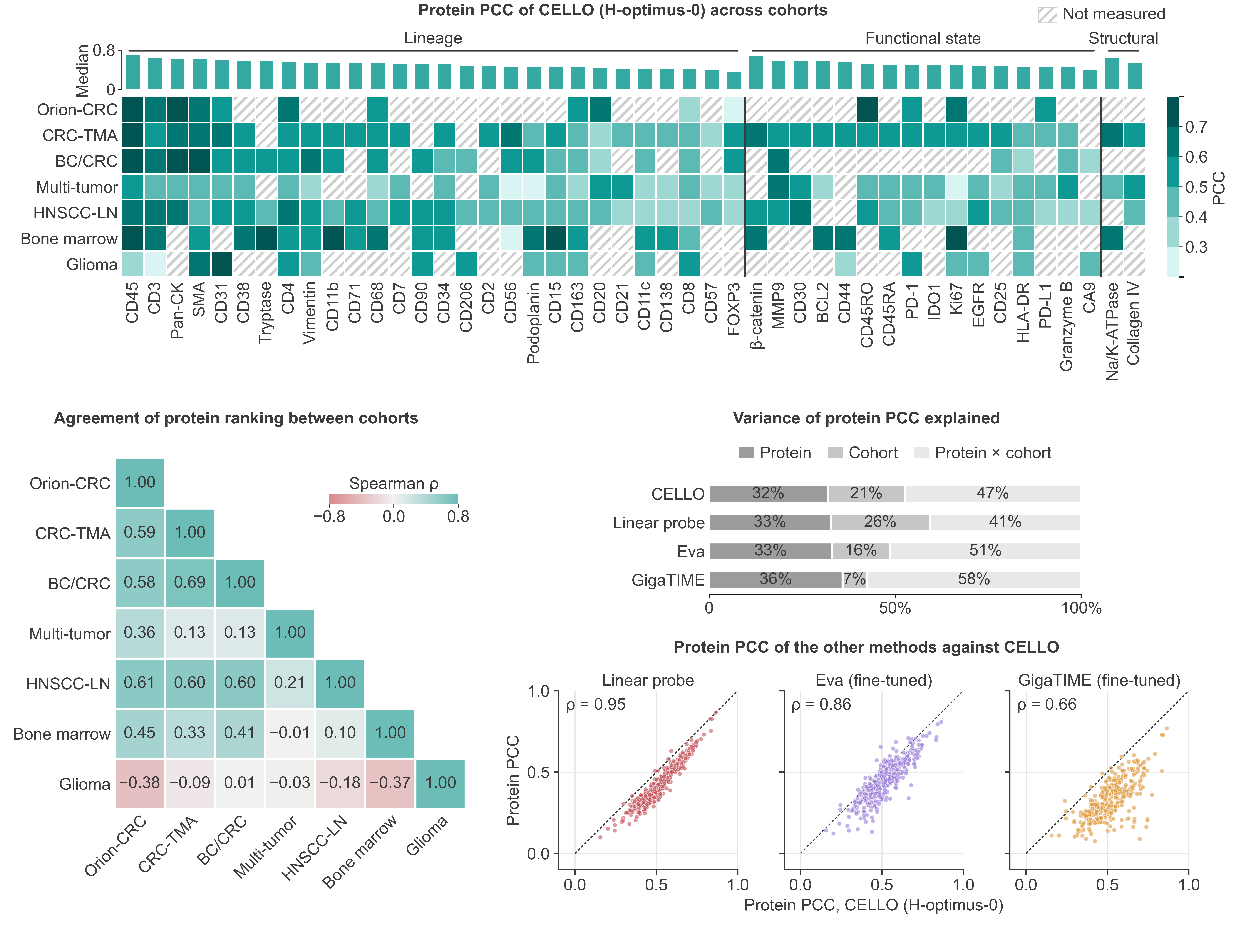}
  \caption{Per-protein accuracy across cohorts. Top: protein PCC of CELLO (H-optimus-0) for every protein measured in at least three cohorts, grouped into lineage, functional-state and structural proteins; bars: median over cohorts; hatched: not measured. Bottom left: Spearman correlation of protein PCC between pairs of cohorts over their shared proteins. Bottom right: share of the variance of protein PCC explained by protein, cohort and their interaction (46 proteins measured in at least three cohorts), and protein PCC of the other methods against CELLO; $\rho$: Spearman correlation.}
  \label{fig:protein_map}
\end{figure*}

\begin{figure*}[t]
  \centering
  \includegraphics[width=\linewidth]{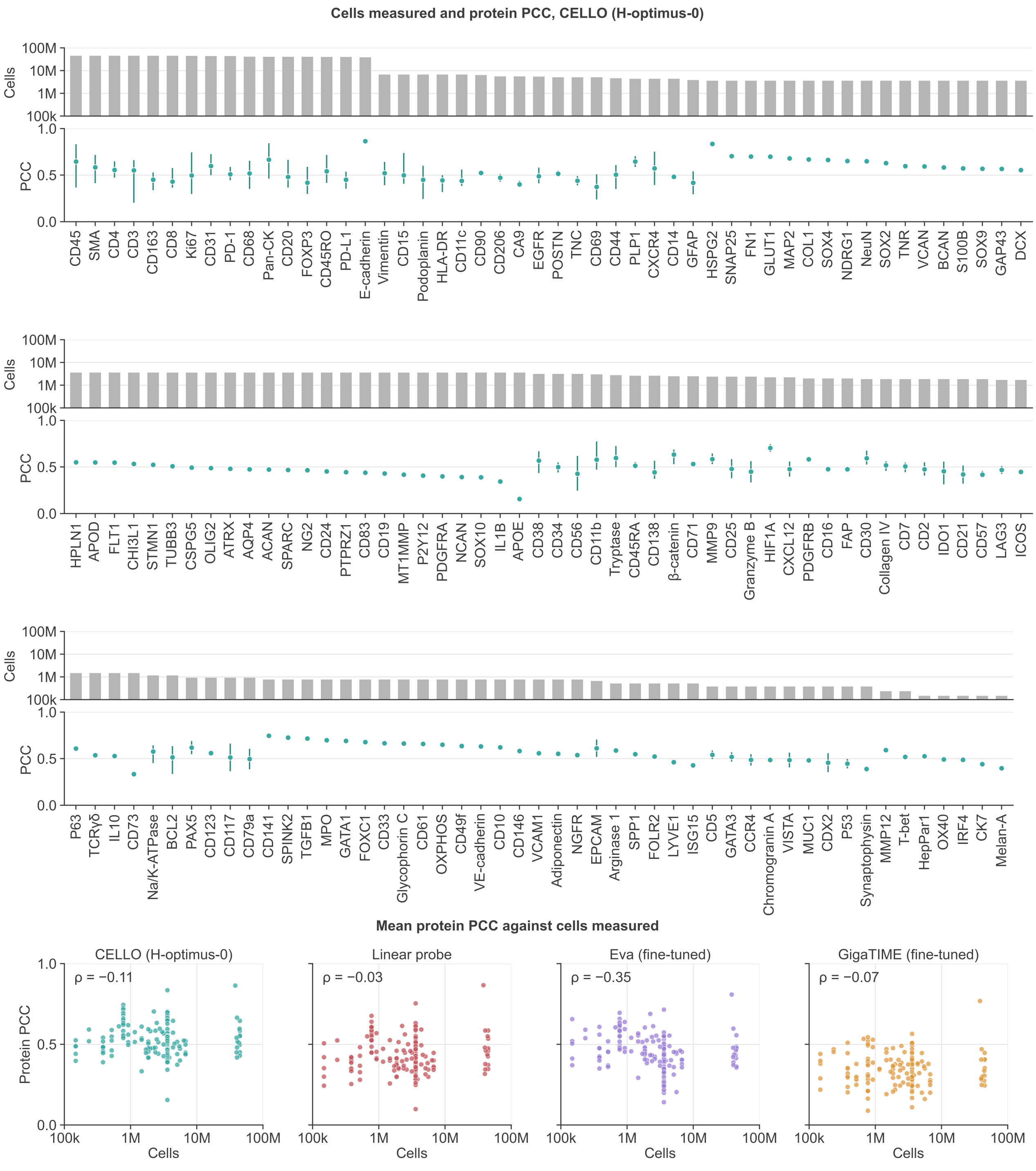}
  \caption{Accuracy against the number of cells measured. Top: for each of the 149 proteins, the number of cells in which it was measured over all cohorts, and the PCC of CELLO (H-optimus-0), mean and range over cohorts. Bottom: mean protein PCC against cells measured for four methods; $\rho$: Spearman correlation.}
  \label{fig:protein_cells}
\end{figure*}

\Cref{fig:protein_map} shows the PCC of CELLO with H-optimus-0 for every protein measured in at least three cohorts, how the ranking of proteins agrees between cohorts and how much of the variance of protein PCC the protein and the cohort explain; \cref{fig:protein_cells} relates the accuracy of every protein to the number of cells in which it was measured (\cref{sec:proteins}), and \crefrange{tab:prot_orion}{tab:prot_glioma} list the protein PCC and SSIM of every model (\cref{sec:supp_scores}).
The quantities of \cref{sec:proteins} are computed as follows.

\paragraph{Protein PCC.}
For a cohort and a protein, the protein PCC is the mean over the cohort's test samples of the PCC between predicted and measured levels across the sample's cells (\cref{sec:splits}).
Markers of different panels are merged by their canonical protein name, which gives 149 proteins and 341 cohort--protein pairs.
The spread within a cohort is the range of protein PCC over its proteins, and a protein counts in favor of CELLO against fine-tuned Eva when its protein PCC is higher.

\paragraph{Variance decomposition.}
For the 46 proteins measured in at least three cohorts (210 cohort--protein pairs), we fit an additive linear model of protein PCC with protein and cohort as categorical factors and compute its type-II sums of squares.
The shares reported are the sums of squares of protein, cohort and residual divided by their total; with one score per cohort--protein pair, the residual of the additive model is the protein-by-cohort interaction.

\paragraph{Agreement between cohorts.}
For each pair of cohorts, we compute the Spearman correlation of protein PCC over the proteins measured in both, 8 to 47 per pair.

\paragraph{Cells measured.}
The number of cells of a protein is the sum of the cells, after quality control, of the cohorts whose panel measures it, and its accuracy is the mean of its protein PCC over these cohorts; the two are compared by Spearman correlation over the 149 proteins.

\paragraph{Protein properties.}
The properties of \cref{fig:props} are computed from the measured levels $y$ of a protein, the targets of \cref{sec:task}, on each test sample and averaged over the test samples of the cohort.
Sparsity is the Gini coefficient of the linear intensity $e^{y} - 1$ over the sample's cells, and dynamic range is the 99th minus the 50th percentile of $y$.
For spatial autocorrelation, $y$ is standardized over the sample's cells to $z$, at most 20{,}000 focal cells $F$ are drawn at random, and Moran's $I = \sum_{i \in F} z_i \bar{z}_{N(i)} / \sum_{i \in F} z_i^2$, where $\bar{z}_{N(i)}$ is the mean of $z$ over the 10 cells nearest to cell $i$.
Subcellular localization is assigned to each protein by its main compartment (nuclear, cytoplasmic, secreted or extracellular matrix, otherwise membrane).
Quintiles are taken over the 341 cohort--protein pairs pooled across cohorts, and the Kruskal--Wallis tests compare protein PCC across the five quintiles or the four localizations.
The margin over fine-tuned Eva is the difference of protein PCC for each pair, and its trend is the Spearman correlation of this margin with Moran's $I$ over the 341 pairs.

\section{Protein maps of the other cohorts}
\label{sec:supp_maps}
\begin{figure*}[t]
  \centering
  \includegraphics[width=\linewidth]{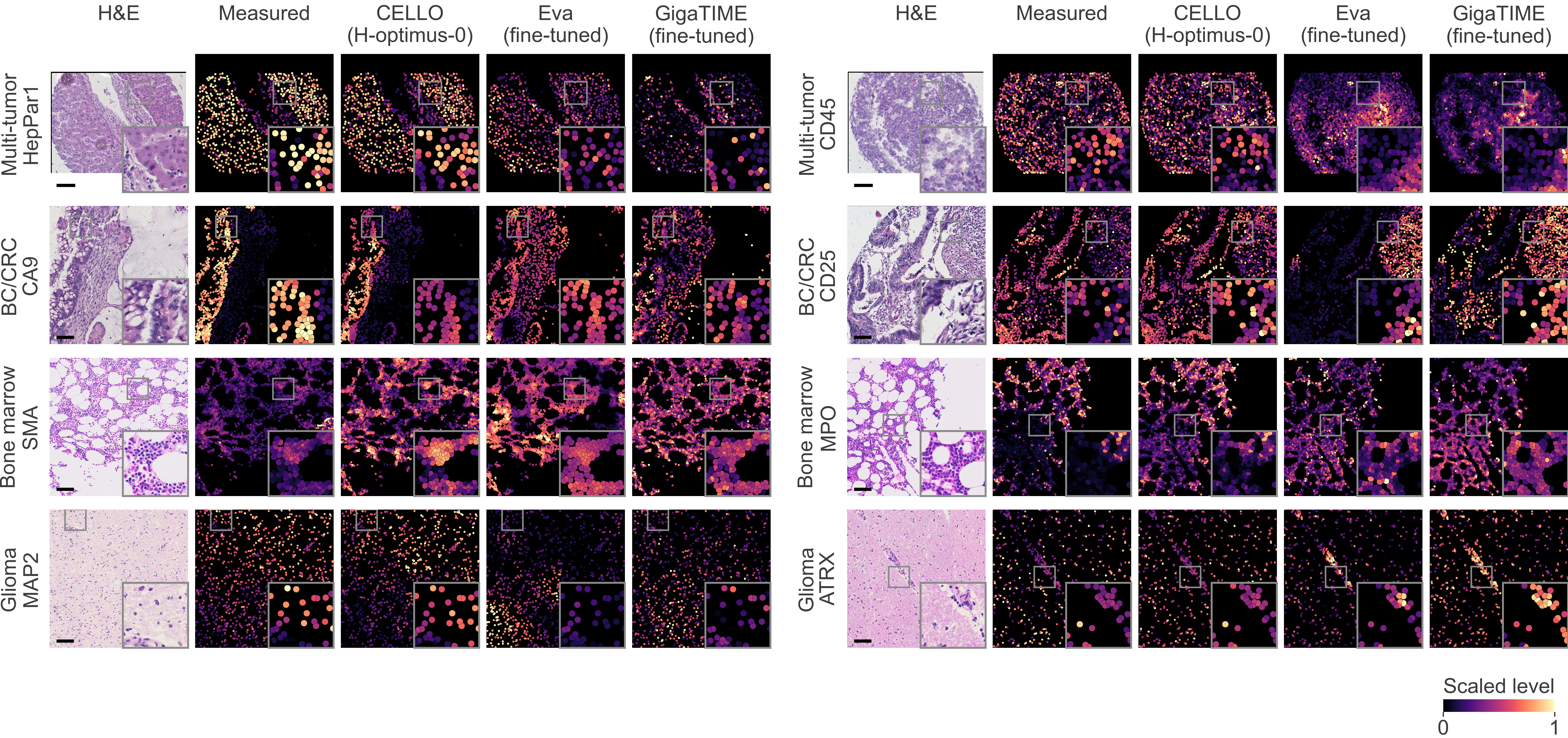}
  \caption{Protein maps on held-out samples of the four cohorts not shown in \cref{fig:maps}, displayed in the same way. Each case shows the H\&E of one view and its cells colored by one protein, as measured and as predicted; the inset magnifies the boxed region; every panel is scaled to the 1st--99th percentile of its cells in the view; scale bar, 100\,$\mu$m.}
  \label{fig:maps_supp}
\end{figure*}

\Cref{fig:maps_supp} shows two cases for each of Multi-tumor (HepPar1, CD45), BC/CRC (CA9, CD25), bone marrow (smooth-muscle actin, MPO) and glioma (MAP2, ATRX).
Each view is an 800\,$\mu$m square of the sample, or the whole core when the core is smaller, and the inset a 120\,$\mu$m square within it.
The correlation $r$ of a method in a view, here and in \cref{sec:qualitative}, is the PCC between its predicted and the measured levels of the shown protein over the cells inside the view.
Over the cells of each view, CELLO with H-optimus-0 reaches $r = 0.65$ to $0.89$, against $-0.36$ to $0.36$ for fine-tuned Eva and $-0.41$ to $0.21$ for fine-tuned GigaTIME.

\section{Per-protein scores}
\label{sec:supp_scores}
\Crefrange{tab:prot_orion}{tab:prot_glioma} give, for each cohort, the PCC and SSIM of every model on every protein of its panel, from the test predictions of \cref{tab:main}.
The score of a protein is the mean over the cohort's test samples of its PCC or SSIM across the cells of a sample; \cref{tab:main} averages the same per-sample scores over the proteins of a sample and then over samples.

\begin{table*}[p]
\centering
\small
\resizebox{\linewidth}{!}{\setlength{\aboverulesep}{0pt}\setlength{\belowrulesep}{0pt}
\begin{tabular}{l|cc|cc|cc|cc|cc|cc|cc|cc|cc|cc}
\toprule
 & \multicolumn{4}{c|}{Eva} & \multicolumn{4}{c|}{GigaTIME} & \multicolumn{6}{c|}{Linear probe} & \multicolumn{6}{c}{CELLO} \\
 & \multicolumn{2}{c|}{zero-shot} & \multicolumn{2}{c|}{fine-tuned} & \multicolumn{2}{c|}{zero-shot} & \multicolumn{2}{c|}{fine-tuned} & \multicolumn{2}{c|}{UNI2-h} & \multicolumn{2}{c|}{Virchow2} & \multicolumn{2}{c|}{H-optimus-0} & \multicolumn{2}{c|}{UNI2-h} & \multicolumn{2}{c|}{Virchow2} & \multicolumn{2}{c}{H-optimus-0} \\
\cmidrule(lr){2-3} \cmidrule(lr){4-5} \cmidrule(lr){6-7} \cmidrule(lr){8-9} \cmidrule(lr){10-11} \cmidrule(lr){12-13} \cmidrule(lr){14-15} \cmidrule(lr){16-17} \cmidrule(lr){18-19} \cmidrule(lr){20-21}
Protein & PCC & SSIM & PCC & SSIM & PCC & SSIM & PCC & SSIM & PCC & SSIM & PCC & SSIM & PCC & SSIM & PCC & SSIM & PCC & SSIM & PCC & SSIM \\
\midrule
E-cadherin & 0.687 & 0.293 & 0.809 & 0.490 & -- & -- & 0.769 & 0.436 & \textbf{0.866} & 0.666 & 0.859 & \textbf{0.702} & \textbf{0.866} & 0.676 & 0.858 & 0.654 & 0.846 & 0.653 & 0.864 & 0.689 \\
Pan-CK & 0.594 & 0.253 & 0.795 & 0.552 & 0.373 & 0.220 & 0.728 & 0.507 & 0.830 & 0.642 & 0.821 & 0.625 & 0.826 & 0.623 & 0.834 & 0.663 & 0.819 & 0.656 & \textbf{0.842} & \textbf{0.675} \\
CD45 & 0.672 & 0.377 & 0.697 & 0.544 & -- & -- & 0.729 & 0.416 & 0.829 & 0.663 & 0.818 & \textbf{0.682} & 0.825 & 0.674 & 0.830 & 0.608 & 0.810 & 0.587 & \textbf{0.832} & 0.612 \\
CD45RO & 0.572 & 0.316 & 0.612 & 0.413 & -- & -- & 0.554 & 0.296 & 0.703 & 0.510 & 0.696 & 0.523 & 0.698 & 0.511 & \textbf{0.723} & \textbf{0.589} & 0.692 & 0.528 & 0.716 & 0.584 \\
SMA & 0.392 & 0.150 & 0.661 & 0.365 & -- & -- & 0.570 & 0.279 & 0.674 & 0.571 & 0.663 & 0.567 & 0.672 & 0.589 & 0.681 & 0.604 & 0.659 & 0.572 & \textbf{0.686} & \textbf{0.624} \\
\bottomrule
\end{tabular}
}
\caption{Per-protein PCC and SSIM on Orion-CRC (16 proteins, 8 test samples): mean over test samples of the protein's score across the cells of a sample. Proteins are sorted by the PCC of CELLO with H-optimus-0. \textbf{Bold}: best per row and metric; --: protein without a matching channel of zero-shot GigaTIME.}
\label{tab:prot_orion}
\end{table*}

\begin{table*}[p]
\centering
\small
\resizebox{\linewidth}{!}{\setlength{\aboverulesep}{0pt}\setlength{\belowrulesep}{0pt}
\begin{tabular}{l|cc|cc|cc|cc|cc|cc|cc|cc|cc|cc}
\toprule
 & \multicolumn{4}{c|}{Eva} & \multicolumn{4}{c|}{GigaTIME} & \multicolumn{6}{c|}{Linear probe} & \multicolumn{6}{c}{CELLO} \\
 & \multicolumn{2}{c|}{zero-shot} & \multicolumn{2}{c|}{fine-tuned} & \multicolumn{2}{c|}{zero-shot} & \multicolumn{2}{c|}{fine-tuned} & \multicolumn{2}{c|}{UNI2-h} & \multicolumn{2}{c|}{Virchow2} & \multicolumn{2}{c|}{H-optimus-0} & \multicolumn{2}{c|}{UNI2-h} & \multicolumn{2}{c|}{Virchow2} & \multicolumn{2}{c}{H-optimus-0} \\
\cmidrule(lr){2-3} \cmidrule(lr){4-5} \cmidrule(lr){6-7} \cmidrule(lr){8-9} \cmidrule(lr){10-11} \cmidrule(lr){12-13} \cmidrule(lr){14-15} \cmidrule(lr){16-17} \cmidrule(lr){18-19} \cmidrule(lr){20-21}
Protein & PCC & SSIM & PCC & SSIM & PCC & SSIM & PCC & SSIM & PCC & SSIM & PCC & SSIM & PCC & SSIM & PCC & SSIM & PCC & SSIM & PCC & SSIM \\
\midrule
CD45 & 0.427 & 0.339 & 0.605 & 0.458 & -- & -- & 0.474 & 0.218 & 0.652 & 0.604 & 0.636 & 0.567 & 0.653 & 0.599 & 0.699 & 0.676 & 0.699 & 0.678 & \textbf{0.709} & \textbf{0.688} \\
$\beta$-catenin & 0.343 & 0.216 & \textbf{0.705} & 0.587 & -- & -- & 0.573 & 0.548 & 0.598 & 0.555 & 0.604 & 0.569 & 0.601 & 0.561 & 0.660 & 0.633 & 0.666 & 0.641 & 0.681 & \textbf{0.658} \\
SMA & 0.360 & 0.218 & 0.528 & 0.261 & -- & -- & 0.350 & 0.173 & 0.569 & 0.509 & 0.545 & 0.477 & 0.569 & 0.510 & 0.624 & 0.586 & 0.619 & 0.583 & \textbf{0.639} & \textbf{0.602} \\
Na/K-ATPase & 0.086 & 0.082 & 0.562 & 0.311 & -- & -- & 0.368 & 0.133 & 0.522 & 0.435 & 0.490 & 0.399 & 0.516 & 0.431 & 0.621 & 0.565 & 0.614 & 0.560 & \textbf{0.633} & \textbf{0.578} \\
CD31 & 0.303 & 0.155 & 0.466 & 0.201 & -- & -- & 0.215 & 0.103 & 0.578 & 0.466 & 0.541 & 0.412 & 0.577 & 0.466 & 0.616 & 0.567 & 0.595 & 0.542 & \textbf{0.630} & \textbf{0.581} \\
\bottomrule
\end{tabular}
}
\caption{Per-protein PCC and SSIM on CRC-TMA (56 proteins, 130 test samples), as in \cref{tab:prot_orion}.}
\label{tab:prot_crc_tma}
\end{table*}

\begin{table*}[p]
\centering
\small
\resizebox{\linewidth}{!}{\setlength{\aboverulesep}{0pt}\setlength{\belowrulesep}{0pt}
\begin{tabular}{l|cc|cc|cc|cc|cc|cc|cc|cc|cc|cc}
\toprule
 & \multicolumn{4}{c|}{Eva} & \multicolumn{4}{c|}{GigaTIME} & \multicolumn{6}{c|}{Linear probe} & \multicolumn{6}{c}{CELLO} \\
 & \multicolumn{2}{c|}{zero-shot} & \multicolumn{2}{c|}{fine-tuned} & \multicolumn{2}{c|}{zero-shot} & \multicolumn{2}{c|}{fine-tuned} & \multicolumn{2}{c|}{UNI2-h} & \multicolumn{2}{c|}{Virchow2} & \multicolumn{2}{c|}{H-optimus-0} & \multicolumn{2}{c|}{UNI2-h} & \multicolumn{2}{c|}{Virchow2} & \multicolumn{2}{c}{H-optimus-0} \\
\cmidrule(lr){2-3} \cmidrule(lr){4-5} \cmidrule(lr){6-7} \cmidrule(lr){8-9} \cmidrule(lr){10-11} \cmidrule(lr){12-13} \cmidrule(lr){14-15} \cmidrule(lr){16-17} \cmidrule(lr){18-19} \cmidrule(lr){20-21}
Protein & PCC & SSIM & PCC & SSIM & PCC & SSIM & PCC & SSIM & PCC & SSIM & PCC & SSIM & PCC & SSIM & PCC & SSIM & PCC & SSIM & PCC & SSIM \\
\midrule
MMP9 & 0.026 & 0.030 & 0.577 & 0.247 & -- & -- & 0.346 & 0.148 & 0.496 & 0.346 & 0.518 & 0.351 & 0.516 & 0.360 & 0.576 & 0.485 & 0.588 & 0.502 & \textbf{0.602} & \textbf{0.518} \\
CD123 & 0.238 & 0.167 & \textbf{0.685} & \textbf{0.578} & -- & -- & 0.595 & 0.570 & 0.469 & 0.452 & 0.504 & 0.482 & 0.491 & 0.469 & 0.555 & 0.541 & 0.578 & 0.559 & 0.586 & 0.575 \\
CD20 & $-$0.021 & 0.028 & \textbf{0.567} & 0.345 & 0.026 & 0.043 & 0.293 & 0.187 & 0.401 & 0.162 & 0.422 & 0.168 & 0.401 & 0.159 & 0.541 & 0.427 & 0.559 & 0.445 & 0.562 & \textbf{0.448} \\
Granzyme B & 0.072 & 0.058 & 0.528 & 0.267 & -- & -- & 0.313 & 0.150 & 0.466 & 0.308 & 0.478 & 0.313 & 0.482 & 0.319 & 0.538 & 0.439 & 0.546 & 0.451 & \textbf{0.562} & \textbf{0.469} \\
Collagen IV & $-$0.007 & 0.004 & 0.518 & 0.243 & -- & -- & 0.338 & 0.142 & 0.455 & 0.331 & 0.473 & 0.336 & 0.475 & 0.346 & 0.533 & 0.447 & 0.546 & 0.461 & \textbf{0.561} & \textbf{0.478} \\
\bottomrule
\end{tabular}
}
\caption{Per-protein PCC and SSIM on Multi-tumor (57 proteins, 66 test samples), as in \cref{tab:prot_orion}.}
\label{tab:prot_multitumor}
\end{table*}

\begin{table*}[p]
\centering
\small
\resizebox{\linewidth}{!}{\setlength{\aboverulesep}{0pt}\setlength{\belowrulesep}{0pt}
\begin{tabular}{l|cc|cc|cc|cc|cc|cc|cc|cc|cc|cc}
\toprule
 & \multicolumn{4}{c|}{Eva} & \multicolumn{4}{c|}{GigaTIME} & \multicolumn{6}{c|}{Linear probe} & \multicolumn{6}{c}{CELLO} \\
 & \multicolumn{2}{c|}{zero-shot} & \multicolumn{2}{c|}{fine-tuned} & \multicolumn{2}{c|}{zero-shot} & \multicolumn{2}{c|}{fine-tuned} & \multicolumn{2}{c|}{UNI2-h} & \multicolumn{2}{c|}{Virchow2} & \multicolumn{2}{c|}{H-optimus-0} & \multicolumn{2}{c|}{UNI2-h} & \multicolumn{2}{c|}{Virchow2} & \multicolumn{2}{c}{H-optimus-0} \\
\cmidrule(lr){2-3} \cmidrule(lr){4-5} \cmidrule(lr){6-7} \cmidrule(lr){8-9} \cmidrule(lr){10-11} \cmidrule(lr){12-13} \cmidrule(lr){14-15} \cmidrule(lr){16-17} \cmidrule(lr){18-19} \cmidrule(lr){20-21}
Protein & PCC & SSIM & PCC & SSIM & PCC & SSIM & PCC & SSIM & PCC & SSIM & PCC & SSIM & PCC & SSIM & PCC & SSIM & PCC & SSIM & PCC & SSIM \\
\midrule
Pan-CK & 0.600 & 0.508 & 0.739 & 0.627 & 0.333 & 0.161 & 0.612 & 0.454 & 0.743 & 0.561 & 0.707 & 0.500 & 0.731 & 0.523 & 0.790 & \textbf{0.722} & 0.782 & 0.706 & \textbf{0.792} & 0.721 \\
CD45 & 0.495 & 0.436 & 0.668 & 0.616 & -- & -- & 0.565 & 0.522 & 0.671 & 0.497 & 0.646 & 0.450 & 0.668 & 0.478 & 0.712 & 0.609 & 0.708 & 0.620 & \textbf{0.726} & \textbf{0.630} \\
SMA & 0.433 & 0.294 & 0.587 & 0.366 & -- & -- & 0.406 & 0.236 & 0.652 & 0.586 & 0.625 & 0.546 & 0.656 & 0.587 & 0.709 & 0.683 & 0.696 & 0.675 & \textbf{0.717} & \textbf{0.692} \\
EPCAM & 0.463 & 0.374 & 0.635 & 0.497 & -- & -- & 0.542 & 0.408 & 0.653 & 0.572 & 0.631 & 0.549 & 0.645 & 0.557 & 0.695 & 0.605 & 0.693 & 0.605 & \textbf{0.704} & \textbf{0.622} \\
CD3 & 0.448 & 0.227 & 0.596 & 0.479 & 0.060 & 0.021 & 0.511 & 0.410 & 0.552 & 0.520 & 0.537 & 0.512 & 0.554 & 0.531 & 0.631 & 0.600 & 0.631 & 0.600 & \textbf{0.655} & \textbf{0.623} \\
\bottomrule
\end{tabular}
}
\caption{Per-protein PCC and SSIM on BC/CRC (36 proteins, 45 test samples), as in \cref{tab:prot_orion}.}
\label{tab:prot_bccrc}
\end{table*}

\begin{table*}[p]
\centering
\small
\resizebox{\linewidth}{!}{\setlength{\aboverulesep}{0pt}\setlength{\belowrulesep}{0pt}
\begin{tabular}{l|cc|cc|cc|cc|cc|cc|cc|cc|cc|cc}
\toprule
 & \multicolumn{4}{c|}{Eva} & \multicolumn{4}{c|}{GigaTIME} & \multicolumn{6}{c|}{Linear probe} & \multicolumn{6}{c}{CELLO} \\
 & \multicolumn{2}{c|}{zero-shot} & \multicolumn{2}{c|}{fine-tuned} & \multicolumn{2}{c|}{zero-shot} & \multicolumn{2}{c|}{fine-tuned} & \multicolumn{2}{c|}{UNI2-h} & \multicolumn{2}{c|}{Virchow2} & \multicolumn{2}{c|}{H-optimus-0} & \multicolumn{2}{c|}{UNI2-h} & \multicolumn{2}{c|}{Virchow2} & \multicolumn{2}{c}{H-optimus-0} \\
\cmidrule(lr){2-3} \cmidrule(lr){4-5} \cmidrule(lr){6-7} \cmidrule(lr){8-9} \cmidrule(lr){10-11} \cmidrule(lr){12-13} \cmidrule(lr){14-15} \cmidrule(lr){16-17} \cmidrule(lr){18-19} \cmidrule(lr){20-21}
Protein & PCC & SSIM & PCC & SSIM & PCC & SSIM & PCC & SSIM & PCC & SSIM & PCC & SSIM & PCC & SSIM & PCC & SSIM & PCC & SSIM & PCC & SSIM \\
\midrule
CD30 & 0.231 & 0.174 & \textbf{0.683} & 0.457 & -- & -- & 0.560 & 0.310 & 0.506 & 0.477 & 0.534 & 0.508 & 0.522 & 0.490 & 0.661 & 0.634 & 0.663 & 0.638 & 0.674 & \textbf{0.648} \\
HIF1A & $-$0.004 & 0.032 & \textbf{0.709} & 0.559 & -- & -- & 0.604 & 0.513 & 0.554 & 0.493 & 0.549 & 0.491 & 0.565 & 0.494 & 0.649 & 0.609 & 0.644 & 0.611 & 0.662 & \textbf{0.627} \\
CD45 & 0.439 & 0.365 & 0.601 & 0.444 & -- & -- & 0.521 & 0.292 & 0.586 & 0.563 & 0.568 & 0.552 & 0.589 & 0.563 & 0.648 & 0.621 & 0.641 & 0.617 & \textbf{0.657} & \textbf{0.632} \\
CD4 & 0.300 & 0.208 & 0.621 & 0.492 & 0.131 & 0.048 & 0.564 & 0.465 & 0.525 & 0.491 & 0.520 & 0.488 & 0.533 & 0.494 & 0.619 & 0.584 & 0.614 & 0.584 & \textbf{0.631} & \textbf{0.600} \\
CD3 & 0.415 & 0.240 & 0.579 & 0.320 & 0.190 & 0.071 & 0.521 & 0.304 & 0.542 & 0.515 & 0.534 & 0.508 & 0.546 & 0.517 & 0.618 & 0.588 & 0.611 & 0.586 & \textbf{0.630} & \textbf{0.601} \\
\bottomrule
\end{tabular}
}
\caption{Per-protein PCC and SSIM on HNSCC-LN (57 proteins, 388 test samples), as in \cref{tab:prot_orion}.}
\label{tab:prot_hnscc}
\end{table*}

\begin{table*}[p]
\centering
\small
\resizebox{\linewidth}{!}{\setlength{\aboverulesep}{0pt}\setlength{\belowrulesep}{0pt}
\begin{tabular}{l|cc|cc|cc|cc|cc|cc|cc|cc|cc|cc}
\toprule
 & \multicolumn{4}{c|}{Eva} & \multicolumn{4}{c|}{GigaTIME} & \multicolumn{6}{c|}{Linear probe} & \multicolumn{6}{c}{CELLO} \\
 & \multicolumn{2}{c|}{zero-shot} & \multicolumn{2}{c|}{fine-tuned} & \multicolumn{2}{c|}{zero-shot} & \multicolumn{2}{c|}{fine-tuned} & \multicolumn{2}{c|}{UNI2-h} & \multicolumn{2}{c|}{Virchow2} & \multicolumn{2}{c|}{H-optimus-0} & \multicolumn{2}{c|}{UNI2-h} & \multicolumn{2}{c|}{Virchow2} & \multicolumn{2}{c}{H-optimus-0} \\
\cmidrule(lr){2-3} \cmidrule(lr){4-5} \cmidrule(lr){6-7} \cmidrule(lr){8-9} \cmidrule(lr){10-11} \cmidrule(lr){12-13} \cmidrule(lr){14-15} \cmidrule(lr){16-17} \cmidrule(lr){18-19} \cmidrule(lr){20-21}
Protein & PCC & SSIM & PCC & SSIM & PCC & SSIM & PCC & SSIM & PCC & SSIM & PCC & SSIM & PCC & SSIM & PCC & SSIM & PCC & SSIM & PCC & SSIM \\
\midrule
CD11b & 0.306 & 0.085 & 0.722 & 0.624 & -- & -- & 0.361 & 0.357 & 0.695 & 0.682 & 0.687 & 0.675 & 0.702 & 0.689 & 0.762 & 0.746 & 0.758 & 0.742 & \textbf{0.774} & \textbf{0.753} \\
CXCR4 & 0.351 & 0.233 & 0.722 & 0.582 & -- & -- & 0.582 & 0.544 & 0.686 & 0.657 & 0.685 & 0.644 & 0.694 & 0.674 & 0.743 & 0.727 & 0.742 & 0.729 & \textbf{0.752} & \textbf{0.738} \\
CD141 & 0.074 & 0.032 & 0.658 & 0.553 & -- & -- & 0.214 & 0.209 & 0.674 & 0.638 & 0.645 & 0.583 & 0.677 & 0.644 & 0.734 & 0.706 & 0.729 & 0.701 & \textbf{0.745} & \textbf{0.718} \\
HIF1A & 0.182 & 0.078 & 0.668 & 0.540 & -- & -- & 0.203 & 0.143 & 0.651 & 0.527 & 0.623 & 0.462 & 0.662 & 0.541 & 0.729 & 0.676 & 0.726 & 0.664 & \textbf{0.745} & \textbf{0.686} \\
Ki67 & 0.128 & 0.023 & 0.668 & 0.426 & $-$0.159 & $-$0.051 & 0.509 & 0.378 & 0.679 & 0.662 & 0.670 & 0.630 & 0.683 & 0.657 & 0.735 & 0.725 & 0.733 & 0.724 & \textbf{0.744} & \textbf{0.733} \\
\bottomrule
\end{tabular}
}
\caption{Per-protein PCC and SSIM on bone marrow (52 proteins, 11 test samples), as in \cref{tab:prot_orion}.}
\label{tab:prot_bm}
\end{table*}

\begin{table*}[p]
\centering
\small
\resizebox{\linewidth}{!}{\setlength{\aboverulesep}{0pt}\setlength{\belowrulesep}{0pt}
\begin{tabular}{l|cc|cc|cc|cc|cc|cc|cc|cc|cc|cc}
\toprule
 & \multicolumn{4}{c|}{Eva} & \multicolumn{4}{c|}{GigaTIME} & \multicolumn{6}{c|}{Linear probe} & \multicolumn{6}{c}{CELLO} \\
 & \multicolumn{2}{c|}{zero-shot} & \multicolumn{2}{c|}{fine-tuned} & \multicolumn{2}{c|}{zero-shot} & \multicolumn{2}{c|}{fine-tuned} & \multicolumn{2}{c|}{UNI2-h} & \multicolumn{2}{c|}{Virchow2} & \multicolumn{2}{c|}{H-optimus-0} & \multicolumn{2}{c|}{UNI2-h} & \multicolumn{2}{c|}{Virchow2} & \multicolumn{2}{c}{H-optimus-0} \\
\cmidrule(lr){2-3} \cmidrule(lr){4-5} \cmidrule(lr){6-7} \cmidrule(lr){8-9} \cmidrule(lr){10-11} \cmidrule(lr){12-13} \cmidrule(lr){14-15} \cmidrule(lr){16-17} \cmidrule(lr){18-19} \cmidrule(lr){20-21}
Protein & PCC & SSIM & PCC & SSIM & PCC & SSIM & PCC & SSIM & PCC & SSIM & PCC & SSIM & PCC & SSIM & PCC & SSIM & PCC & SSIM & PCC & SSIM \\
\midrule
HSPG2 & 0.255 & 0.093 & 0.715 & 0.195 & -- & -- & 0.507 & 0.135 & 0.753 & 0.681 & 0.724 & 0.594 & 0.754 & 0.667 & 0.832 & \textbf{0.817} & 0.822 & 0.809 & \textbf{0.835} & 0.813 \\
CD31 & 0.243 & 0.031 & 0.596 & 0.233 & -- & -- & 0.482 & 0.185 & 0.643 & 0.625 & 0.630 & 0.596 & 0.649 & 0.632 & 0.723 & 0.718 & 0.706 & 0.696 & \textbf{0.727} & \textbf{0.724} \\
SNAP25 & 0.021 & 0.037 & 0.531 & 0.314 & -- & -- & 0.396 & 0.265 & 0.615 & 0.606 & 0.624 & 0.622 & 0.609 & 0.597 & 0.702 & 0.674 & 0.638 & 0.628 & \textbf{0.703} & \textbf{0.688} \\
FN1 & 0.232 & 0.126 & 0.554 & 0.154 & -- & -- & 0.443 & 0.113 & 0.612 & 0.564 & 0.604 & 0.540 & 0.619 & 0.575 & 0.688 & 0.666 & 0.692 & 0.674 & \textbf{0.699} & \textbf{0.679} \\
GLUT1 & 0.189 & 0.068 & 0.549 & 0.165 & -- & -- & 0.420 & 0.119 & 0.623 & 0.561 & 0.605 & 0.543 & 0.632 & 0.582 & \textbf{0.704} & \textbf{0.686} & 0.696 & 0.681 & 0.697 & 0.685 \\
\bottomrule
\end{tabular}
}
\caption{Per-protein PCC and SSIM on glioma (68 markers, 9 test samples), as in \cref{tab:prot_orion}; DCX and DCXv2 are two markers of the panel for the same protein.}
\label{tab:prot_glioma}
\end{table*}

\end{document}